\documentclass[twoside]{article}
\usepackage[accepted]{aistats2021}

\usepackage[round]{natbib}
\renewcommand{\bibname}{References}
\renewcommand{\bibsection}{\subsubsection*{\bibname}}

\usepackage{microtype}
\usepackage{graphicx}
\usepackage{amsmath}
\usepackage{amsthm}
\usepackage{thm-restate}
\usepackage{amsfonts}
\usepackage{booktabs}
\usepackage{rotating}
\usepackage{multicol}
\usepackage{xcolor}
\usepackage{multirow}
\usepackage{caption}
\usepackage{subcaption}
\usepackage{alltt}
\usepackage[hidelinks]{hyperref}
\usepackage{url}
\usepackage{algorithm}
\usepackage{algorithmic}
\usepackage{bm}
\usepackage{cleveref}

\usepackage[round]{natbib}
\renewcommand{\bibname}{References}
\renewcommand{\bibsection}{\subsubsection*{\bibname}}
\newcommand{\code}[1]{\mbox{\texttt{#1}}}

\usepackage{xcolor}

\DeclareMathOperator*{\argmax}{arg\,max}
\DeclareMathOperator*{\argmin}{arg\,min}
\DeclareMathOperator*{\E}{\mathbb{E}}
\newcommand\given[1][]{\:#1\vert\:}
\DeclareMathOperator{\imone}{IM1}
\DeclareMathOperator{\imtwo}{IM2}
\DeclareMathOperator{\pd}{p_{\mathcal{D}}}

\newcommand{\norm}[1]{\left\lVert#1\right\rVert}

\begin{document}

\runningtitle{Interpretable Counterfactual Explanations}

\runningauthor{Schut*, Key*, McGrath, Costabello, Sacaleanu, Corcoran, Gal}

\twocolumn[

\aistatstitle{Generating Interpretable Counterfactual Explanations By Implicit Minimisation of Epistemic and Aleatoric Uncertainties}

\aistatsauthor{ Lisa Schut{$^{*\dagger}$}  \And Oscar Key{$^{*\dagger}$}  \And  Rory McGrath{$^\ddagger$} }

\aistatsauthor{Luca Costabello{$^\ddagger$} \And  Bogdan Sacaleanu{$^\ddagger$} \And Medb Corcoran{$^\ddagger$} \And Yarin Gal{$^\dagger$} }
\vspace{0.2cm}
\aistatsaddress{{$^*$}Equal contribution, correspondence to \code{\{schut@robots.ox, oscar.key.20@ucl\}.ac.uk}\\  {$^\dagger$}OATML, University of Oxford, \ {$^\ddagger$}Accenture Labs}
]

\begin{abstract}

Counterfactual explanations (CEs) are a practical tool for demonstrating \emph{why} machine learning classifiers make particular decisions.
For CEs to be useful, it is important that they are easy for users to interpret.
Existing methods for generating interpretable CEs rely on auxiliary generative models, which may not be suitable for complex datasets, and incur engineering overhead.
We introduce a simple and fast method for generating interpretable CEs in a white-box setting without an auxiliary model, by using the predictive uncertainty of the classifier.
Our experiments show that our proposed algorithm generates more interpretable CEs, according to IM1 scores \citep{van2019interpretable}, than existing methods.
Additionally, our approach allows us to estimate the uncertainty of a CE, which may be important in safety-critical applications, such as those in the medical domain. 

\end{abstract}

\section{INTRODUCTION} \label{sec:introduction}
The growing number of decisions influenced by machine learning models drives the need for explanations of \emph{why} a system makes a particular prediction \citep{gdpr}.
Explanations are necessary for users to understand \textit{what factors} influence a decision and understand \textit{what changes} they could make to alter it. One important application for such explanations is \textit{recourse}, where the explanations allow users to understand what adjustments they could make to the input to change the classification given by the model \citep{spangher2018actionable}

A common approach is to generate a counterfactual explanation (CE) of the form ``\textit{If X had not occurred, then Y would not have occurred}'' \citep{wachter2017counterfactual}. Consider the following binary classification problem: ``\textit{Given the current specifications of my house (e.g., location, number of bedrooms, etc.), am I likely or unlikely to sell it for \$300,000?}''. On inputting the details of their apartment, the user might receive the classification ``unlikely''. In this example, a CE could be the same house with upgraded furnishings to increase the desirability, resulting in the classification ``likely''.

Methods for generating CEs focus on finding an alternate input that is \textit{close} to the original input, but with the desired classification \citep{molnar2019}. However, this highlights a fundamental difficulty in designing CEs, namely their similarity to adversarial examples. Both CEs and adversarial examples search for a minimal perturbation to add to the original input that changes the classification. The distinguishing conceptual feature is interpretability:  while CEs should be interpretable, adversarial examples need not be \footnote{Although there is a common conception that adversarial attacks generate \textit{imperceptible changes}, the term `perceptible' is ill-defined, and many adversarial perturbations are visible to the human eye \citep[see e.g.][]{papernot2016limitations, sharif2016accessorize, brown2017adversarial}.} However, interpretability is an ambiguous term, with varying definitions in existing literature \citep{lipton2018mythos}.

\begin{figure}[h]
    \center
    \includegraphics[ width = 0.232 \linewidth, keepaspectratio]{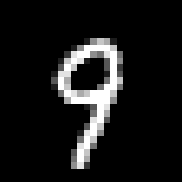}
    \includegraphics[ width = 0.232 \linewidth, keepaspectratio]{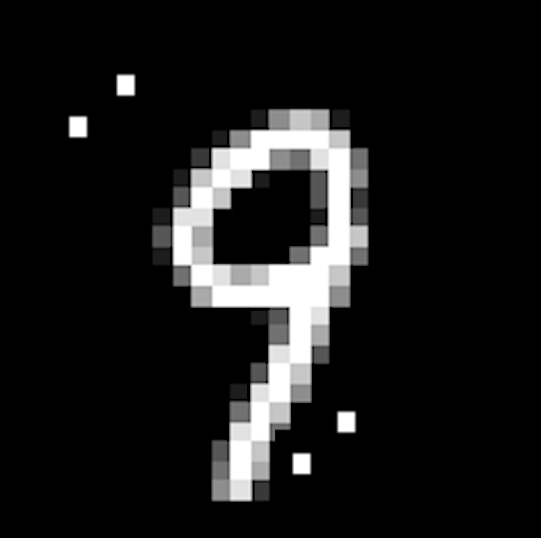}
    \includegraphics[ width = 0.232 \linewidth, keepaspectratio]{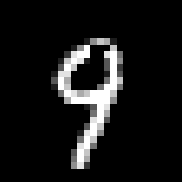}
    \includegraphics[ width = 0.232 \linewidth, keepaspectratio]{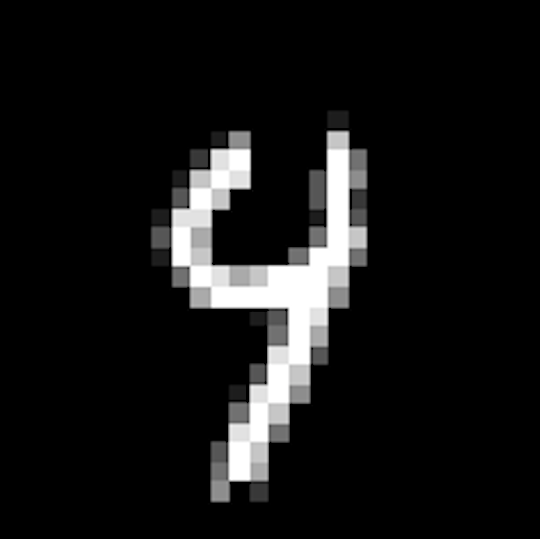}
    \begin{minipage}{0.232\linewidth}
\begin{center}
    Original
\end{center}
\end{minipage}
\begin{minipage}{0.232\linewidth}
\begin{center}
    Unrealistic CE
\end{center}
\end{minipage}
\begin{minipage}{0.232\linewidth}
\begin{center}
    Ambiguous CE
\end{center}
\end{minipage}
\begin{minipage}{0.232\linewidth}
\begin{center}
    Unambiguous, Realistic CE
\end{center}
\end{minipage}
    \captionof{figure}{
        Examples of possible CEs for an input image of the digit $9$ when changing the classification to $4$.
        From left to right: the original image, an unrealistic CE, an ambiguous CE (it can be interpreted as either a $4$ or $9$), a realistic and unambiguous CE (output from our algorithm).
    }
    \label{fig:unamb_realistic}
\end{figure}

We propose defining an interpretable CE as one that is \textbf{realistic}, i.e., a likely scenario for the  user in question, and \textbf{unambiguous}, i.e., not a pathological `borderline' case. \Cref{fig:unamb_realistic} provides an illustration of these two properties for an MNIST image \citep{lecun2010mnist}.
Here, we want to find a minimal change to to alter the original image of a $9$ so that it is $4$.
Second from the left is an example of a CE that is not realistic -- it doesn't resemble a ``normal" $4$. Third from the left is an example of an ambiguous counterfactual; it is unclear whether it depicts a $4$ or $9$. In the final image is a CE that is both realistic and unambiguous, which is clearly preferable.
We give a more extensive definition of realism and unambiguity in \Cref{sec:desiderata}. 

Existing work largely focuses on generating realistic CEs, and does not consider ambiguity \citep{wachter2017counterfactual, dhurandhar2018explanations, joshi2019towards}.
Additionally, many of these approaches rely on using an auxiliary generative model, in addition to the classifier, to either generate realistic CEs or evaluate the realism of CEs in order to guide a search process.
This may impose a bottleneck, as generative models are ill-suited for some datasets, and  incur engineering and maintenance overhead.

In this work, we propose capturing realism and ambiguity using the predictive uncertainty of the classifier.
We consider two types of uncertainty: epistemic and aleatoric uncertainty \citep{kendall2017uncertainties}.
Epistemic uncertainty is uncertainty due to a lack of knowledge, stemming from observing only a subset of all possible data points.
We propose that CEs for which the classifier has low epistemic uncertainty are more realistic, because they are more likely under the data distribution.
Aleatoric uncertainty captures inherent stochasticity in the dataset, for example due to points that lie on the decision boundary between two classes.
Therefore, CEs with lower aleatoric uncertainty will have lower ambiguity.
In \Cref{sec:uncertainty_method} we discuss both concepts in more depth.

Based on these insights, we introduce a novel method for generating interpretable CEs by using a classifier that offers estimates of epistemic and aleatoric uncertainty.
This method does not require an auxiliary generative model and requires less hyperparameter tuning than existing methods.
Existing neural network classifiers can be easily extended to represent uncertainty, for example, by using Monte Carlo dropout \citep{gal2016dropout}, thus this approach has a low engineering cost.
Additionally, for many applications where it is necessary to offer an explanation, it may also be essential to quantify the uncertainty in the predictions.
Thus, uncertainty estimates might already be available and could readily be used for generating CEs.

Our contributions are that we:
\begin{itemize}
    \item link the concepts of aleatoric and epistemic uncertainty to the concepts of unambiguous and realistic CEs (\Cref{sec:uncertainty_method}),
    \item introduce a new method for generating interpretable CEs based on implicit minimisation of both epistemic and aleatoric uncertainty (\Cref{sec:uncertainty_method}),
    \item demonstrate empirically, from both a qualitative and quantitative perspective, that our method generates more interpretable CEs than existing methods, despite not requiring an auxiliary model (\Cref{sec:eval}).
\end{itemize}

We release an implementation of our algorithm, and the experiments, at \href{https://github.com/oscarkey/explanations-by-minimizing-uncertainty}{github.com/oscarkey/explanations-by-minimizing-uncertainty}.

\section{CE DESIDERATA}\label{sec:desiderata}
In this section we define the desirable properties of CEs, including those which make a CE interpretable.

Before doing this, we clarify the term `counterfactual explanation'.
Consider an initial input $x$ which is to be explained.
We can write the alternative input, $x'$, found as the explanation as $x' = x + \Delta$, where $\Delta$ is the minimal change.
From here on we will use counterfactual \emph{explanation} to refer to $x'$, and counterfactual \emph{perturbation} (CP) to refer to $\Delta$.

Explanation desiderata are subjective, and some are not mentioned below. Our goal is not to define a complete list of all possible desiderata, but simply to make explicit the framework and targets we consider in this work.
If interested, the reader can refer to \citet{lipton2018mythos} for a more in depth discussion.

For each desideratum below, we illustrate it using the example given in the introduction: a landlord has a two bedroom, one bathroom, one garage house in Boston with a small garden.
A classifier answers the question ``Is this property likely to sell for \$300,000?'' with \texttt{False}.
The goal is to generate explanations of the form ``\textit{If the property had X, then the classifier would return \texttt{True}}''.

\paragraph{Minimal Perturbation}
The CE should be as similar as possible to the original instance, i.e. there should be as few changes as possible between $x$ and $x'$ \citep{huysmans2011empirical, wachter2017counterfactual, molnar2019, laugel2019dangers, van2019interpretable}.
By making as few changes as possible, we produce concise explanations that are more interpretable and avoid information overload \citep{lahav2018interpretable}.
For example, consider the following two CPs that both change the classification of the aforementioned problem to \texttt{True}:
\begin{itemize}
\item [$\Delta_1$ ] repainting the kitchen
\item [$\Delta_2$ ] repainting both the kitchen and the bathroom
\end{itemize}
As both obtain the desired outcome, $\Delta_1$ is more desirable as it is more concise.

\paragraph{Realistic Explanation}
The suggested explanation must be from a ``possible world'' \citep{wachter2017counterfactual}. This is important because the explanation must represent a concept that the user understands in order for it to be informative to them. For example, the explanation ``\textit{if the garage was rebuilt into $100$ small rooms, then it is likely the house could be sold for $\$300$,$000$}'' is clearly unrealistic and not informative to the user. In comparison, ``\textit{if the garage was rebuilt into an ensuite bedroom, then it is likely the house could be sold for $\$300$,$000$}'' would be a reasonable explanation. In addition, the feature values must be realistic when considered \textit{together} \citep{joshi2019towards}. For example, a one-bedroom house with $6$ bathrooms would not be a realistic explanation as most real houses have a higher bedroom to bathroom ratio.

\paragraph{Unambiguous Explanation}
CEs should be unambiguous to be \textit{informative}.
In this context, we take informative to mean explanations that humans can understand and learn from.
For example, doctors may be interested in informative explanations from a breast cancer detection model.

Ambiguous inputs may be classified with a low confidence score, result in  `borderline' cases or inputs that resemble multiple classifications.
For example, an 'ambiguous' house specification is one that one buyer might value over $\$300$,$000$, but another buyer might value under $\$300$,$000$.
For a visual example, see \Cref{fig:unamb_realistic}, where the input resembles both a $4$ and $9$.

\paragraph{Realistic or Actionable Perturbation}
It must be possible for the user to apply the suggested CP in practice.
While the `Realistic Explanation' property ensures that the \emph{explanation} is a possible instance, it will only provide the user with recourse if it is possible for them to apply the suggested \emph{perturbation} to transition from their original input to the explanation.
For example, while having an identical house to the original but in New York city would be a realistic counterfactual, it is not an actionable perturbation because the user cannot move their house to a different city.

\paragraph{Run Time of the Algorithm}
The algorithm must generate CEs sufficiently quickly for the use case \citep{van2019interpretable}.
While other computational properties of the algorithm, such as memory usage, are also important, we highlight run time because recourse is often offered in a user facing application, so the algorithm must be able to generate CEs sufficiently quickly for this interactive setting.
Many generation algorithms involve non-convex optimisation and repeated evaluations of a potentially expensive model, thus run time is a significant concern.

In our approach, we will target all of the above desiderata. We explicitly target the desiderata \textit{unambiguous} and \textit{realistic}, through our design of the loss function. We believe these desiderata are particularly important as they distinguish CEs from adversarial examples. The remaining desiderata are targeted implicitly through the design of the optimisation procedure of our CE generation algorithm. In the next section, we will introduce our method and discuss how each desideratum is addressed.

\section{METHODOLOGY} \label{sec:uncertainty_method}

In this section we introduce a method for generating interpretable CEs.
In particular, we introduce and motivate using epistemic and aleatoric uncertainty to capture realism and unambiguity.
Next, we show that minimizing both types of uncertainty can be implemented efficiently by minimizing the cross-entropy objective of specific model classes.
Based on these insights, we present a fast, greedy algorithm that generates minimal perturbations that minimize both types of uncertainty, resulting in interpretable explanations.
Note that our method is a post-hoc -- this method is used on trained classifiers to generate CEs.

\subsection{Uncertainty as a Proxy for Realism and Unambiguity}
We begin by following \citet{wachter2017counterfactual} in framing the task of generating CEs as an optimisation problem.
Given an input $x$, we can generate an explanation $x'$ in class $y'$ by solving
\begin{equation} \label{eq:objective_with_h}
    x' = \argmin_{x'} \max_{\lambda \leq \Lambda } \lambda \ell(f,x',y') + h(x'),
\end{equation}
where $f$ is the classifier, $\ell(\cdot)$ is a loss function, $\Lambda$ is a hyperparameter, and $h(\cdot)$ is a measure of intepretability (for which lower is better).
Intuitively, we want to generate an explanation in class $y'$, which is encouraged by $\ell(\cdot)$, and is interpretable, as encouraged by $h(\cdot)$.
The main difficulty is the definition of $h(\cdot)$, the measure of interpretability.
As previously introduced, we define $h(\cdot)$ by considering two key aspects of interpretability: realism and unambiguity.

First we consider how to generate \emph{realistic} CEs.
As we discuss in \Cref{sec:related-work}, existing literature has revealed that this property is the most difficult to achieve, thus improving on it is the primary focus our work.
We suggest that, when generating a CE $x'$ in target class $y'$, we should maximise $\pd(x' \given y')$, where $\pd$ is the training data distribution.
Our justification for this builds on the work of \citet{dhurandhar2018explanations, joshi2019towards, van2019interpretable}, as we discuss in detail in \Cref{sec:related-work-generating}.
In short, explanations which are likely under the distribution of the training data will appear familiar to the user and thus realistic.
Specifically, we should consider the distribution for the target class, i.e. $\pd(x' \given y')$, in order to generate examples which look realistic for the particular target class $y'$.
For example, it would not be realistic for a house classified as expensive to be very small and in a cheap area.

Given this definition of \emph{realistic}, Bayes' rule gives us the following expression for the un-normalized density,
\begin{align}
    p_{\mathcal{D}}(x' | y') &= \frac{p_{\mathcal{D}}(y' | x') p_{\mathcal{D}}(x')}{p_{\mathcal{D}}(y')} \\
    &\propto p_{\mathcal{D}}(y' | x') p_{\mathcal{D}}(x') \label{eq:unorm-density}.
\end{align}
If we use a standard classification model with a softmax output, then $p_{\mathcal{D}}(y'|x')$ is estimated by the output of the model.
To compute $p_{\mathcal{D}}(x')$, the likelihood of $x'$ under the training data distribution, we have several choices. One option would be to use a separate generative model to estimate $p_{\mathcal{D}}(x')$. This would lead us to a similar objective to that introduced by \citet{dhurandhar2018explanations} and \citet{van2019interpretable}.

Instead, we note that we can approximate $p_{\mathcal{D}}(x')$ without the need for an additional model by using a classifier that offers estimates of uncertainty over its predictions \citep{smith2018understanding, grathwohl2019your}.
In particular, we can use the estimate of \emph{epistemic uncertainty}. This is uncertainty about which function is most suitable to explain the data, because there are many possible functions which fit the finite training data available.
Considering the input space, a Bayesian classifier will have lower epistemic uncertainty on points which are close to the training data, and the uncertainty will increase as we move away from it. Thus epistemic uncertainty should be negatively correlated with $p_{\mathcal{D}}(x')$. \citet{gal2018sufficient} show empirically that this is in fact the case for Bayesian neural networks implemented using deep ensembles. Thus, given a classifier which offers estimates of epistemic uncertainty we can compute an un-normalized value for $p_{\mathcal{D}}(x' | y')$.

Second, we consider how to generate \textit{unambiguous} CEs. To capture ambiguity, we use \textit{aleatoric uncertainty}.
This type of uncertainty arises due to inherent noisiness, or stochasticity, in the data distribution \citep{smith2018understanding}.
To generate unambiguous CEs, we generate explanations in areas of the input space where the classifier has low aleatoric uncertainty.

\subsection{Uncertainty in Practice}
There are several different approaches for obtaining classifiers that offer estimates of epistemic and aleatoric uncertainty.
For the experiments in this paper we choose to use an ensemble of deep neural networks, as this is a simple method for computing high quality uncertainty estimates \citep{lakshminarayanan2017simple}.
Contrary to other methods for estimating uncertainty in deep learning, deep ensembles place no constraints on the architecture class of the classifier. Note that our approach will work with any model that offers uncertainty estimates.

We define $h(x')$ as the predictive entropy of the classifier when evaluated on input $x'$.
Predictive entropy captures both aleatoric and epistemic uncertainty, and both are low when the predictive entropy is low.
Specifically, the predictive entropy estimated using ensembles is
\begin{align} \label{eq:uncertainty}
    h(x') &= - \sum_{y \in \mathcal{Y}} \Bar{p}(y|x') \log \Bar{p}(y|x') \\
    \Bar{p}(y|x') &= \frac{1}{M} \sum_m p_m(y|x'),
\end{align}
where we have $M$ models in the ensemble \citep{smith2018understanding}.
Here, $p_m(y|x')$ is the softmax output of the $m$th model in the ensemble.

Having defined $h(x')$ as the predictive entropy, we note that the term $\lambda h(x')$ in \Cref{eq:objective_with_h} is redundant.
This is because a counterfactual that minimizes the cross-entropy (i.e., that maximizes the probability to be assigned to a class) must also minimize the predictive entropy (i.e., be likely under our approximation of the data distribution).
Formally,
\begin{restatable}{proposition}{propsameminimum}
\label{prop:same_minimum}
    For a classification model $f$, $\argmin_{x'} \ell(f,x',y') \in \argmin_{x'} h(x')$, where $l(\cdot)$ is cross-entropy and $h(\cdot)$ is predictive entropy.
\end{restatable}
We provide a formal derivation in \Cref{proofs}. The intuition behind this proposition is that the cross-entropy is minimized (equal to $0$) when the target class has a probability $1$ and all other classes have probability $0$. In this scenario, predictive entropy will also be minimized at $0$.

As a result, we drop the term $\lambda h(\cdot)$ from \Cref{eq:objective_with_h} and objective becomes simply
\begin{equation} \label{eq:objective_without_h}
    x' = \argmin_{x'} \ell(f,x',y').
\end{equation}

This simplification of the objective makes it cheaper and easier to generate CEs.
We avoid the minimax optimization of the parameter $\lambda$, which might otherwise increase the computational cost of the optimization.
Additionally, it eliminates the hyperparameter $\Lambda$, which would otherwise need to be tuned.
As we discuss in detail in \Cref{sec:related-work-generating}, this is an improvement over existing approaches such as \citet{wachter2017counterfactual}, which requires both minimax optimization and tuning of $\Lambda$, and \citet{van2019interpretable}, which uses an objective with several hyperparameters.

\subsection{A Greedy Algorithm for Generating Minimal CEs} \label{sec:jsma}
Above, we propose an objective function for generating realistic CEs, and show that it can be implicitly minimized by minimizing the cross-entropy loss.
If we optimize the loss function directly, we will generate a sample in class $y'$.
However, this does not incorporate the \textit{minimality} or \textit{realistic perturbation} properties.
Our approach to ensuring both of these properties are satisfied is to constrain the optimization process through the optimization algorithm.
Specifically, we extend the Jacobian-based Saliency Map Attack (JSMA), originally introduced by \citet{papernot2016limitations} for the purpose of generating adversarial examples.
We adapt this algorithm to generate meaningful perturbations.

JSMA is an iterative algorithm that updates the most salient feature, i.e. the feature that has the largest influence on the classification, by $\delta$ at each step.
To generate realistic CEs rather than adversarial examples, we replace the original definition of saliency by defining the most salient feature as that which has the largest gradient with respect to the objective in \Cref{eq:objective_without_h}:
\begin{equation} \label{eq:most_salient_feature}
    \operatorname{most\_salient\_feature}(x') = \argmax_j \nabla_{x_j'} \left[ \ell (f,x', y') \right],
\end{equation}
where $\nabla$ denotes the partial derivative and $x_j'$ denotes the $j$th feature of $x$.
Updating each feature iteratively by $\delta$ acts as a heuristic for minimising the $L_0$ distance between the original input and the CE \citep{papernot2016limitations}.
The algorithm terminates when the input is classified as the target class with \textit{high confidence}, or after reaching the maximum number of iterations.
Alternatively, the algorithm can be configured to fail if the explanation does not reach the predefined confidence level.
This enforces that generated explanations are those on which the classifier has low uncertainty, which may be important for certain applications.
We give pseudocode in \Cref{algorithm_counterfactuals_intext}.

This is a fast algorithm for generating realistic and unambiguous explanations using minimal perturbations.
We also want to ensure that the perturbation is realistic and actionable.
In many cases, we can manually identify the features a user cannot change, and lock these features to prevent the algorithm from perturbing them.
For example, we might prevent the algorithm from changing the location of a house.
This simple approach assumes that we are explicitly aware of factors that can be changed, which is often but not always the case.
We leave a detailed investigation into other approaches for generating realistic perturbations for future work.

\begin{algorithm}[t]
\caption{Generating Counterfactuals\\ (For more detail, see \Cref{appx:cf_alg})}
\begin{algorithmic}[1] \label{algorithm_counterfactuals_intext}
    \STATE \textbf{Input} original observation $x$; target class $y'$; ensemble of models $ \{f_m \}_{m=1}^M$; maximum number of iterations $N$; minimum confidence of target class $\gamma$; perturbation size $\delta$; maximum number of times each feature is changed $n$; optional: a function that clips the values to a permitted pre-defined range clip.
    \STATE \textbf{Output} counterfactual $x'$
    \STATE $x' \leftarrow x$
    \STATE $c \leftarrow 0$
    \STATE $P = 0_{n_p}$, where $n_p$ is the number of input features
    \WHILE{$\frac{1}{M} \sum_m p_m(y'| x') \leq \gamma$ and $c \leq N$}
    \STATE Compute forward derivative \\  $S(x',y') = \nabla_{x'} \frac{1}{M} \sum_m^M {\ell}(f_m, x', y')$
    \STATE $i = \text{argmax}_{i:i \in P, P[i]<n} S(x',y')[i]$
    \STATE $x'[i] = x'[i] + \text{sign}(S(x',y')[i]) \cdot \delta$
    \STATE $x'= \text{clip}(x')$
    \STATE $P[i] \leftarrow P[i] + 1$
    \STATE $c \leftarrow c+1$
    \ENDWHILE
    \STATE \textbf{return} $x'$
\end{algorithmic}
\end{algorithm}

\subsection{Adversarial Training }

Our proposed method works with any classifier that both offers uncertainty estimates and for which we have access to the gradients (of cross-entropy loss with respect to some input).
However, if it is possible to retrain the classifier then the realism of the generated explanations can be improved by applying adversarial training, as we demonstrate empirically in \Cref{sec:eval}.
Specifically, we augment the dataset during training using adversarial examples generated by FGSM \citep{goodfellow2014explaining}, see \Cref{appx:adv_training} for details.

We suggest that adversarial training might improve the realism of the generated CEs for two reasons.
First, \citet{lakshminarayanan2017simple} demonstrate that adversarial training improves uncertainty estimation, both on in-distribution and out-of-distribution inputs.
This should improve the performance of our method, as we generate CEs in areas of input space where the classifier has low uncertainty.

\begin{figure}
    \centering
    \includegraphics[trim={3.5cm 1.5cm 3.5cm 1.5cm}, clip, width= 0.32\linewidth]{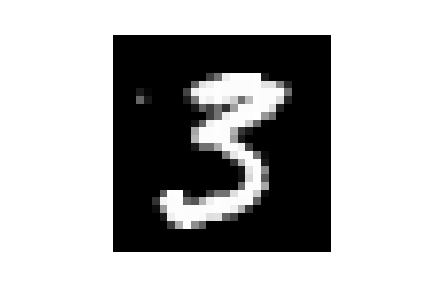}
    \includegraphics[trim={3.5cm 1.5cm 3.5cm 1.5cm}, clip, width= 0.32\linewidth]{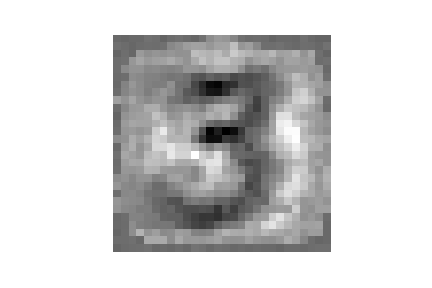}
    \includegraphics[trim={3.5cm 1.5cm 3.5cm 1.5cm}, clip, width= 0.32\linewidth]{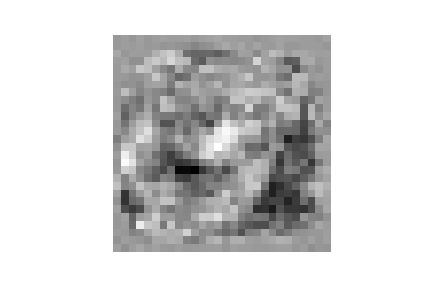}
    \caption{
        Gradients of classifiers trained with adversarial training (middle image), versus without (right image) for an input image (left image).
        We observe that adversarial training results in gradients (which can used to identify a salient features) that are more aligned with human perception.
        This example is inspired by \citet[Figure~2]{tsipras2018robustness}.
    }
    \label{fig:adv_training}
\end{figure}

Second, adversarial training can lead to learning more robust features \citep{tsipras2018robustness, ilyas2019adversarial}.
Augmenting the training set with adversarial examples during training ensures that the model does not focus on noise when learning features for classification.
As such, the model is more likely to learn features that are not noise, and therefore are more interpretable \citep{tsipras2018robustness}.
An example of the effect of adversarial training is shown in \Cref{fig:adv_training}.
The saliency of an adversarially trained model, as shown by the middle image, is more aligned with human interpretation than the saliency of a regular model (shown by the right image).
We discuss these two effects further in \Cref{appx:adv_training}.

\section{RELATED WORK} \label{sec:related-work}

\subsection{Generating CEs} \label{sec:related-work-generating}
Below we summarize the different methods used to generate CEs.
We begin with \citet{wachter2017counterfactual}, who frame the task of finding a CE $x'$ in target class $y'$ for initial input $x$ as the optimization problem
\begin{equation} \label{eq:wachter_objective}
    x' = \argmin_{x'} \max_{\lambda \leq \Lambda} \lambda \ell(f, x', y') + d(x, x'),
\end{equation}
where $f$ is the classifier, $\ell(\cdot)$ is a loss function (the authors use MSE loss), and $d$ is some measure of distance (the authors use a weighted $L_1$ distance), and $\Lambda$ is a hyperparameter.
This is equivalent to the objective function we use in our approach, as given in \Cref{eq:objective_with_h}, if $h(x')$ is defined as distance to the original input.
In comparison to our approach, this definition of $h(x')$ does not give any consideration to ensuring that $x'$ is realistic, and \citet{wachter2017counterfactual} note that it risks generating adversarial examples.

Various approaches adapt \Cref{eq:wachter_objective} in an attempt to generate realistic CEs:
\begin{itemize}
    \item \citet{dhurandhar2018explanations} include an additional penalty in the objective to encourage CEs to lie on the training data manifold.
    The authors fit an auxiliary autoencoder model to the training data.
    In the objective, they then include the reconstruction loss of applying this autoencoder to the CE.
    The assumption is that reconstruction loss will be higher for CEs which are not likely under the training distribution, which will encourage the approach to generate realistic CEs.
    
    \item \citet{van2019interpretable} note that the approach introduced by \citet{dhurandhar2018explanations} does not take into account the data distribution of each class, for example a very large house is unlikely to also be very cheap.
    Thus, the authors include an additional loss term which guides the search process towards a `prototype' instance of the relevant class.
    The prototype for each class is defined as the average location in a latent space of all the training points in that class.
    Again, the authors use an autoencoder to map inputs into the latent space.
    One disadvantage of this approach is that the objective function contains several terms, and a hyperparameter must be specified for each in order to scale them appropriately.
    
    \item The methods above generate CEs by searching in input space. In contrast, \citet{joshi2019towards} search in a latent space, and use a generative model to map instances from this latent space into the input space in order to evaluate them. The objective is
    \begin{equation}
        x' = \argmin_{z \sim p(z)} \min_{\lambda \leq \Lambda} \lambda \ell(f(\mathcal{G}(x')), y') +  c(x, x'),
    \end{equation}
    where $p(z)$ is a distribution over the latent space, $\mathcal{G}(\cdot)$ is the generative model mapping the latent space to input space, and $c(\cdot)$ is a cost function (which we can view the same as the distance function).
    One limitation of this approach is that the CEs are produced by the generative model, thus suffer from the pathologies of that model.
    For example, a VAE is likely to generate blurry explanations.
\end{itemize}
We claim that our approach has several advantages over these methods.
First, we avoid the engineering, and potential computational, overheads of implementing, training, and maintaining an auxiliary generative model.
Second, we have a simple objective function which does not involve minimax optimization or specifying hyperparameters, both of which incur additional computational cost.

The weakness of our method is that it requires a classifier which offers uncertainty estimates.
This is likely to have higher computational cost, and in particular the ensemble of classifiers that we use in our experimental work is more expensive to train and evaluate than a single model.
However, our method can be used with any classifier that offers both epistemic and aleatoric uncertainty, and several fast approaches are available for deep learning models \citep{gal2016dropout, liu2020simpleuncertainty, van2020uncertainty}.
Additionally, we argue that many applications where the machine learning system must offer the user recourse, estimates of the uncertainty in the classification will also be required, and so will already be available from the classifier.
For example, when using a machine learning tool to make a decision, estimates of the uncertainty in the predictions are very important to be able to act cautiously, or defer to a human expert, when the model is unsure.

\subsection{Adversarial examples}
Counterfactual examples are closely related to adversarial examples. Adversarial examples are crafted by finding the minimal perturbations required to change the classification of an image \cite{szegedy2013intriguing}. Mathematically, this can be formulated as
\begin{align}
    \label{eq:adv_example}
    \begin{split}
        \min_{x'} \quad &  d(x',x) \\
        \textrm{s.t.} \quad & f(x') = y',
    \end{split}
\end{align}
where $x$ is the original input, $x'$ is the adversarial example, $y'$ is the target class, $d(\cdot)$ is a distance metric and $f(\cdot)$ is the classifier.

This is very similar to the mathematical formulation used to generate CEs in \Cref{eq:wachter_objective}. In literature, the distinguishing feature between the two fields is \textit{interpretability}. While we want counterfactual examples to be interpretable, adversarial examples need not be. Our work focuses on this distinguishing feature; we design an algorithm that leverages uncertainty to generate interpretable CEs.

\subsection{Evaluation metrics} \label{sec:eval}
To  evaluate CE  generation  algorithms, we evaluate the realism and minimality of the CEs generated.  To measure \emph{minimality}, we report the $L_1$ distance between the original input and the explanation.
Realism is more difficult to measure because it is poorly defined.
In the literature there are several approaches:

\paragraph{Human evaluation}
\cite{dhurandhar2018explanations} use subject experts to evaluate the CEs generated by their approach by hand. This provides ground-truth data on human interpretability. We choose not to use this approach because it is not automated, and so difficult to perform at scale and not suitable for frequent evaluation when tuning hyperparameters.

\paragraph{Vulnerability evaluation \citep{laugel2019dangers}}
This approach is based on the concept of justification: a CE is justified if there is a path in input space between the CE and a point in the training set that does not cross the decision boundary between classes.
The authors introduce an algorithm which evaluates what fraction of the CEs generated by an algorithm are justified.
We choose not to use this approach because the algorithm does not scale to high dimensional input spaces, as it relies on populating an epsilon ball around the explanation with points.
Additionally, it is not clear if the notion of justification relates to the same definition of human interpretability as we use in this work.

\paragraph{IM1 / IM2 \citep{van2019interpretable}}
Two metrics based on the reconstruction losses of autoencoders are
\begin{align}
    \imone(x', y, y') &= \frac{\norm{x' - \mathrm{AE}_{y'}(x')}^2_2}{\norm{x' - \mathrm{AE}_{y}(x')}^2_2 + \epsilon} \\
    \imtwo(x', y') &= \frac{\norm{\mathrm{AE}_{y}(x') - \mathrm{AE}(x')}_2^2}{\norm{x'}_1 + \epsilon}
\end{align}
where $\mathrm{AE}_y$ is an autoencoder trained only on instances from class $y$, and $\mathrm{AE}$ is an autoencoder trained on instances from all classes.
We can see that $\imone$ is the ratio of the reconstruction loss of an autoencoder trained on the counterfactual class divided by the loss of an autoencoder trained on all classes.
$\imtwo$ is the normalized difference between the reconstruction of the CE under an autoencoder trained on the counterfactual class, and one trained on all classes.

We choose to evaluate the realism of the explanations generated by our method using $\imone$.
We omit IM2 because it fails to pass a sanity check. In particular, we find that IM2 scores are not significantly different for out-of-distribution data (i.e., `junk data') than in-distribution data. See Appendix \ref{appx:eval_im2} for further details.

\section{EMPIRICAL ANALYSIS}
\Cref{appx:experiment_setup} gives full details of the configuration we use in each experiment.

\subsection{Datasets}
We perform our analysis on three datasets: MNIST \citep{lecun2010mnist}, the Breast Cancer Wisconsin Diagnostic dataset \citep{Dua2019}, and the Boston Housing dataset \citep{Dua2019}.
We choose MNIST as it is easy to visualize, which allows non-experts to evaluate the interpretability of the generated CEs.
We consider the two tabular datasets because this type of data is frequently used in the interpretability literature\footnote{We could not find one consistently used benchmark; a similar conclusion is drawn by \citet{verma2020counterfactual}.}.

\paragraph{MNIST dataset}
This dataset contains gray-scale images of handwritten digits ranging between $0$ and $9$. The goal of the CE explanation is to find a perturbation that changes the image classification from, e.g., a $1$ to a $7$. We consider MNIST, as image-based data allows us to visually inspect the quality of the CEs. Our classifiers obtain an accuracy of $ 98.5\%$ on the test set.

\paragraph{Breast Cancer Wisconsin Diagnostic dataset}
A tabular dataset where each row contains various measurements of a cell sample from a tumour, alongside a binary diagnosis of whether the tumour is benign or malignant.
A CE for a particular input changes the classification from benign to malignant, or vice-versa.
Our classifiers obtain an accuracy of $96.9\%$ on the validation set.

\paragraph{Boston Housing dataset}
A tabular dataset where each row contains statistics about a suburb of Boston, alongside the median house value.
To construct a classification problem, we divide the dataset into suburbs where the price is below the median, and those where it is above.
A CE for a particular input changes the classification from below the median to above, or vice-versa.
Our classifiers obtain an accuracy of $86.3\%$ on the validation set.

\subsection{Compared Methods}
We benchmark the performance of our method against \citet{van2019interpretable}, a state-of-the-art approach for generating CEs.
We also compare against JSMA, the adversarial attack from which we draw inspiration for our algorithm.
JSMA provides a baseline for interpretatability.
We include JSMA for two reasons: [1] to determine that we are able to generate more interpretable counterfactuals,  and [2] to validate that JSMA can efficiently create minimal perturbations.

\subsection{Results}

\paragraph{Comparison to other approaches}
In \autoref{table:results} we compare our method to \cite{van2019interpretable} and JSMA, reporting both the realism of the CEs and the size of the perturbation.
We find that our approach generates more realistic CEs than \cite{van2019interpretable}, despite not requiring an auxiliary generative model, as can be seen from the lower IM1 scores.

\begin{table}[h]
    \begin{center}
    \captionof{table}{Realism (IM1) and minimality ($L_1$) of generated explanations.
    IM1 is a proxy for the distance to the class data manifold, by using reconstruction errors (see Section \ref{sec:eval}).
    A lower value is better.
    The $L_1$ distance is that between the original input and the explanation.
    We compute the mean of each metric for $100$ randomly selected test points.
    We report the mean of this mean over $10$ seeds, and in brackets the standard deviation over the seeds.
    See \Cref{appx:experiment_setup} for details.
    \textsc{VLK} is the method introduced by \citet{van2019interpretable}.
    Note, the reported scores for VLK are based on our experiments and differ from those reported in their paper.
    We improve on their reported results for MNIST, but find worse performance for the Breast Cancer dataset.
    We discuss the steps we took to reproduce their results in Appendix \ref{appx:exp_details_wbc}.
    }
    \vspace{5pt}
    \begin{small}
    \begin{sc}
    \begin{tabular}{l c c c c} \toprule
    \textbf{Method} & \multicolumn{2}{c}{\textbf{Realism} (IM1)} & \multicolumn{2}{c}{\textbf{Minimality} ($L_1$)} \\
    & mean & std & mean & std \\ \midrule

    \multicolumn{5}{c}{\textit{MNIST}} \\ \hline
    JSMA & $1.11$ & ($0.01$) & ${14.4}$ & ($2.3$) \\
    VLK  & $1.12$ & ($0.06$) &  $47.7$  & ($4.9$) \\
    Ours & ${0.98}$ & ($0.02$) & $38.3$ & ($3.5$) \\ \midrule

    \multicolumn{5}{c}{\textit{Breast Cancer Diagnosis}} \\ \midrule
    JSMA & $1.06$ & ($0.01$) & ${1.23}$ & ($0.04$) \\
    VLK  & $2.81$ & ($0.27$) & $1.27$ & ($0.09$) \\
    Ours & ${0.89}$ & ($0.04$) & $2.57$ & ($0.13$) \\ \midrule

    \multicolumn{5}{c}{\textit{Boston Housing}} \\ \midrule
    JSMA & $1.50$ & ($0.00$) & $0.72$ & ($0.01$) \\
    VLK  & $2.55$ & ($0.08$) & $0.73$ & ($0.20$) \\
    Ours & ${0.85}$ & ($0.00$) & ${1.47}$ & ($0.02$) \\ \bottomrule

    \end{tabular}
    \end{sc}
    \end{small}
    \label{table:results}
    \end{center}
\end{table}

\begin{figure}
\begin{center}
\begin{minipage}{0.233\linewidth}
\begin{center}
    Original
\end{center}
\end{minipage}
\begin{minipage}{0.233\linewidth}
\begin{center}
    Ours
\end{center}
\end{minipage}
\begin{minipage}{0.233\linewidth}
\begin{center}
    JSMA
\end{center}
\end{minipage}
\begin{minipage}{0.233\linewidth}
\begin{center}
    VLK
\end{center}
\end{minipage}
\includegraphics[ width = 0.232 \linewidth, keepaspectratio]{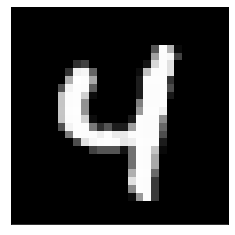}
\includegraphics[ width = 0.232 \linewidth, keepaspectratio]{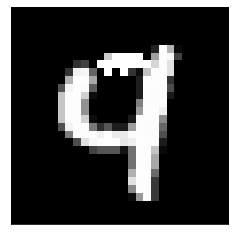}
\includegraphics[ width = 0.232 \linewidth, keepaspectratio]{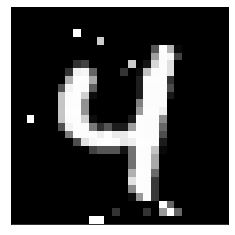}
\includegraphics[ width = 0.232 \linewidth, keepaspectratio]{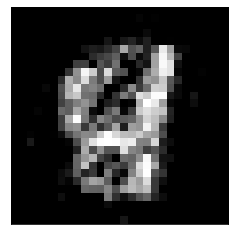}

\includegraphics[ width = 0.232 \linewidth, keepaspectratio]{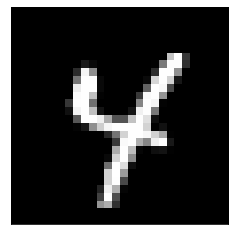}
\includegraphics[ width = 0.232 \linewidth, keepaspectratio]{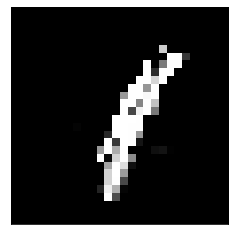}
\includegraphics[ width = 0.232 \linewidth, keepaspectratio]{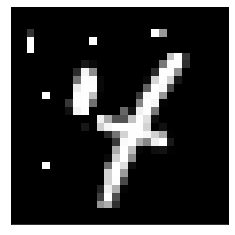}
\includegraphics[ width = 0.232 \linewidth, keepaspectratio]{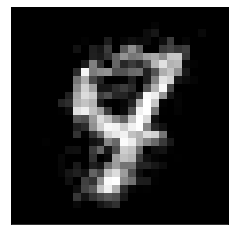}

\includegraphics[ width = 0.232 \linewidth, keepaspectratio]{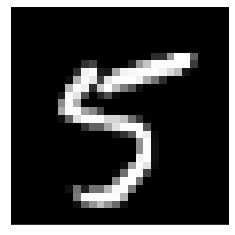}
\includegraphics[ width = 0.232 \linewidth, keepaspectratio]{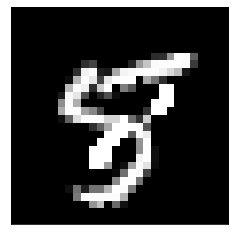}
\includegraphics[ width = 0.232 \linewidth, keepaspectratio]{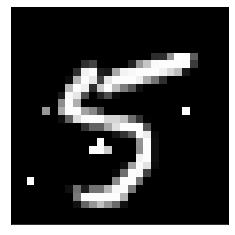}
\includegraphics[ width = 0.232 \linewidth, keepaspectratio]{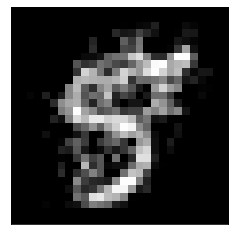}
    \caption{Qualitative Examples of Counterfactual Explanations. Each row shows a different example. From top to bottom, the goal is to change the classification from a $4 \rightarrow 9$, $4 \rightarrow 1$, and $5 \rightarrow 8$.
    The left column shows original images.
    The following three columns show examples of generated CEs by the algorithms we consider.
    From left to right: our algorithm, JSMA, VLK.
    Our algorithm is able to create more realistic counterfactuals -- i.e., images that are likely under the data distribution.
    We note that the CEs above generated by VLK appear less realistic than those shown in the original paper. This is likely because we consider \textit{targeted} CEs, i.e., we randomly specify a target class for the explanation, whereas in \cite{van2019interpretable} the explanation can be in any class, which is an easier task.
    }
    \label{fig:qual_examples}
    \vspace{-0.5cm}
\end{center}
\end{figure}

Comparing our method to JSMA, we note that our method generates larger perturbations but with better IM1 scores.
This demonstrates that our adapted loss function is successfully trading off the size of the perturbation for the realism of the explanation, as desired.
JSMA is able to obtain the lowest $L_1$ as it is an adversarial attack designed to generate minimal perturbations.
We emphasise that we only report the $L_1$ distance of JSMA to show that it can efficiently create minimal perturbations, and it does not generate realistic explanations.
This can be seen in \Cref{fig:qual_examples}, which shows qualitative examples of the explanations generated by the three methods.

\Cref{fig:qual_examples} also shows one failure mode of our proposed algorithm: the strokes in the counterfactuals are less smooth than the strokes in real images. This is due to the algorithm design, which changes single pixels iteratively. Our model does not capture stylistic properties, which would be important if we want to employ our method as a generative model. However, our model is able to grasp high-level changes required, such as adding a white stroke to $4$ to turn it into a $9$. This aspect is more important for explanatory purposes.
In \Cref{appx:add_fig}, we show more examples of CEs generated by our method on both MNIST and tabular data, and provide further insight into which features are altered.

\subsection{Ablation Study}
\begin{figure}
\begin{center}
\includegraphics[ width =0.69\linewidth]{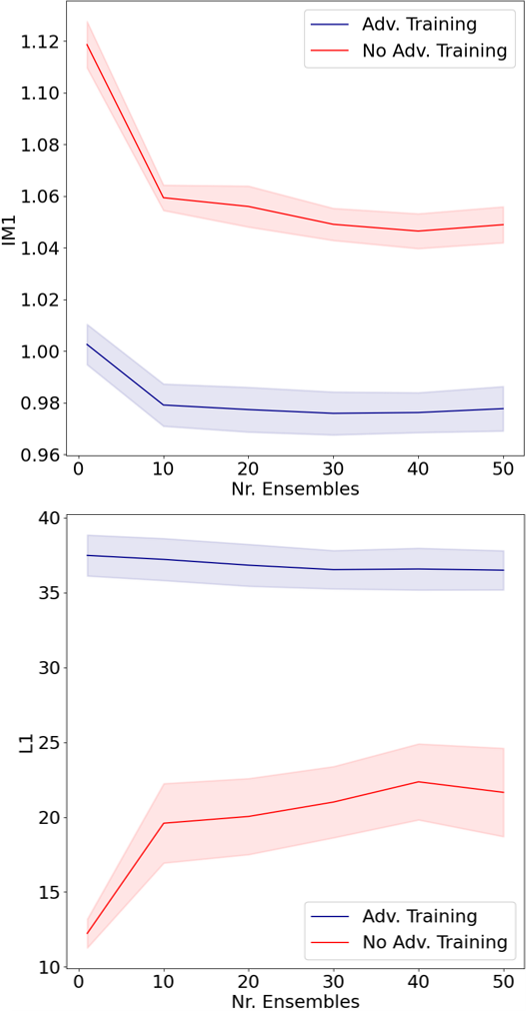}
\caption{
        Ablation Study on the MNIST: effect of adversarial training and number of models in the ensemble on interpretability (top), as measured by the IM1 score, and minimality (bottom), as measure by the mean $L_1$ distance.
        The shaded areas show $95 \%$ CIs, as estimated over $10$ seeds.
    }
    \label{fig:ablation}
\end{center}
\end{figure}

Next, we perform an ablation study to investigate the effects of adversarial training, and the number of models in the ensemble, on the quality of generated CEs for MNIST images. The results are shown in \Cref{fig:ablation}.
Initially, the interpretability of CEs tends to improve as the number of ensemble components increases -- this can be seen from the initial downward slopes of IM1 in the top graph of \Cref{fig:ablation}. However, after $10$ ensembles, the improvement in performance saturates; likely because uncertainty estimation does not improve further. Adversarial training improves the interpretability scores, however leads to less sparse explanations. 

\section{CONCLUSION}
We have introduced a fast method for generating realistic, unambiguous, and minimal CEs. In the process, we collect, define, and discuss the properties which CEs should have.
In comparison to existing methods, our algorithm does not rely on an auxiliary generative model, reducing the engineering overhead.
Nevertheless, we demonstrate empirically that our approach is able to match or exceed the performance of existing methods, with respect to the realism of the CEs generated.
In future work, methodological developments could be explored by adapting the proposed method to work for black-box models \citep{afrabandpey2020decision}.

\newpage 
\subsubsection*{Acknowledgements}
We would like to thank Mizu Nishikawa-Toomey, Andrew Jesson, and Jan Brauner, as well as other members of OATML, for their feedback on the paper. 
Lisa Schut and Oscar Key acknowledge funding from Accenture Labs.

\bibliography{references.bib}

\begin{thebibliography}{}

\bibitem[Afrabandpey et~al., 2020]{afrabandpey2020decision}
Afrabandpey, H., Peltola, T., Piironen, J., Vehtari, A., and Kaski, S. (2020).
\newblock A decision-theoretic approach for model interpretability in bayesian
  framework.
\newblock {\em Machine Learning}, pages 1--22.

\bibitem[Brown et~al., 2017]{brown2017adversarial}
Brown, T.~B., Man{\'e}, D., Roy, A., Abadi, M., and Gilmer, J. (2017).
\newblock Adversarial patch.
\newblock {\em arXiv preprint arXiv:1712.09665}.

\bibitem[Chalasani et~al., 2020]{chalasani2018concise}
Chalasani, P., Chen, J., Chowdhury, A.~R., Jha, S., and Wu, X. (2020).
\newblock Concise explanations of neural networks using adversarial training.
\newblock {\em International Conference on Machine Learning 2020}, pages
  arXiv--1810.06583v9.

\bibitem[Dhurandhar et~al., 2018]{dhurandhar2018explanations}
Dhurandhar, A., Chen, P.-Y., Luss, R., Tu, C.-C., Ting, P., Shanmugam, K., and
  Das, P. (2018).
\newblock Explanations based on the missing: Towards contrastive explanations
  with pertinent negatives.
\newblock In {\em Advances in Neural Information Processing Systems}, pages
  592--603.

\bibitem[Dua and Graff, 2017]{Dua2019}
Dua, D. and Graff, C. (2017).
\newblock {UCI} machine learning repository.

\bibitem[Gal and Ghahramani, 2016]{gal2016dropout}
Gal, Y. and Ghahramani, Z. (2016).
\newblock Dropout as a bayesian approximation: Representing model uncertainty
  in deep learning.
\newblock In {\em International Conference on Machine Learning}, pages
  1050--1059.

\bibitem[Gal and Smith, 2018]{gal2018sufficient}
Gal, Y. and Smith, L. (2018).
\newblock Sufficient conditions for idealised models to have no adversarial
  examples: a theoretical and empirical study with bayesian neural networks.
\newblock {\em arXiv preprint arXiv:1806.00667}.

\bibitem[Goodfellow et~al., 2015]{goodfellow2014explaining}
Goodfellow, I., Shlens, J., and Szegedy, C. (2015).
\newblock Explaining and harnessing adversarial examples.
\newblock In {\em International Conference on Learning Representations}.

\bibitem[Grathwohl et~al., 2020]{grathwohl2019your}
Grathwohl, W., Wang, K.-C., Jacobsen, J.-H., Duvenaud, D., Norouzi, M., and
  Swersky, K. (2020).
\newblock Your classifier is secretly an energy based model and you should
  treat it like one.
\newblock In {\em International Conference on Learning Representations}.

\bibitem[Huysmans et~al., 2011]{huysmans2011empirical}
Huysmans, J., Dejaeger, K., Mues, C., Vanthienen, J., and Baesens, B. (2011).
\newblock An empirical evaluation of the comprehensibility of decision table,
  tree and rule based predictive models.
\newblock {\em Decision Support Systems}, 51(1):141--154.

\bibitem[Ilyas et~al., 2019]{ilyas2019adversarial}
Ilyas, A., Santurkar, S., Tsipras, D., Engstrom, L., Tran, B., and Madry, A.
  (2019).
\newblock Adversarial examples are not bugs, they are features.
\newblock In {\em Advances in Neural Information Processing Systems}, pages
  125--136.

\bibitem[Joshi et~al., 2019]{joshi2019towards}
Joshi, S., Koyejo, O., Vijitbenjaronk, W., Kim, B., and Ghosh, J. (2019).
\newblock Towards realistic individual recourse and actionable explanations in
  black-box decision making systems.
\newblock {\em arXiv preprint arXiv:1907.09615}.

\bibitem[Kendall and Gal, 2017]{kendall2017uncertainties}
Kendall, A. and Gal, Y. (2017).
\newblock {What uncertainties do we need in bayesian deep learning for computer
  vision?}
\newblock In {\em Advances in Neural Information Processing Systems}, pages
  5574--5584.

\bibitem[Kingma and Ba, 2015]{kingma2014adam}
Kingma, D.~P. and Ba, J. (2015).
\newblock Adam: A method for stochastic optimization.
\newblock In {\em ICLR (Poster)}.

\bibitem[Klaise et~al., 2020]{alibi}
Klaise, J., Van~Looveren, A., Vacanti, G., and Coca, A. (2020).
\newblock Alibi: Algorithms for monitoring and explaining machine learning
  models.
\newblock {\em GitHub}.

\bibitem[Lahav et~al., 2018]{lahav2018interpretable}
Lahav, O., Mastronarde, N., and van~der Schaar, M. (2018).
\newblock What is interpretable? using machine learning to design interpretable
  decision-support systems.
\newblock {\em arXiv preprint arXiv:1811.10799}.

\bibitem[Lakshminarayanan et~al., 2017]{lakshminarayanan2017simple}
Lakshminarayanan, B., Pritzel, A., and Blundell, C. (2017).
\newblock Simple and scalable predictive uncertainty estimation using deep
  ensembles.
\newblock In {\em Advances in neural information processing systems}, pages
  6402--6413.

\bibitem[Laugel et~al., 2019]{laugel2019dangers}
Laugel, T., Lesot, M.-J., Marsala, C., Renard, X., and Detyniecki, M. (2019).
\newblock The dangers of post-hoc interpretability: Unjustified counterfactual
  explanations.
\newblock In {\em Proceedings of the Twenty-Eighth International Joint
  Conference on Artificial Intelligence, {IJCAI-19}}, pages 2801--2807.
  International Joint Conferences on Artificial Intelligence Organization.

\bibitem[LeCun et~al., 2010]{lecun2010mnist}
LeCun, Y., Cortes, C., and Burges, C. (2010).
\newblock Mnist handwritten digit database.
\newblock {\em ATT Labs [Online]. Available: http://yann.lecun.com/exdb/mnist},
  2.

\bibitem[Lipton, 2018]{lipton2018mythos}
Lipton, Z.~C. (2018).
\newblock The mythos of model interpretability.
\newblock {\em Queue}, 16(3):31--57.

\bibitem[Liu et~al., 2020]{liu2020simpleuncertainty}
Liu, J., Lin, Z., Padhy, S., Tran, D., Bedrax~Weiss, T., and Lakshminarayanan,
  B. (2020).
\newblock Simple and principled uncertainty estimation with deterministic deep
  learning via distance awareness.
\newblock In Larochelle, H., Ranzato, M., Hadsell, R., Balcan, M.~F., and Lin,
  H., editors, {\em Advances in Neural Information Processing Systems},
  volume~33, pages 7498--7512. Curran Associates, Inc.

\bibitem[Molnar, 2019]{molnar2019}
Molnar, C. (2019).
\newblock {\em Interpretable Machine Learning}.
\newblock \url{https://christophm.github.io/interpretable-ml-book/}.

\bibitem[Papernot et~al., 2016]{papernot2016limitations}
Papernot, N., McDaniel, P., Jha, S., Fredrikson, M., Celik, Z.~B., and Swami,
  A. (2016).
\newblock The limitations of deep learning in adversarial settings.
\newblock In {\em 2016 IEEE European symposium on security and privacy
  (EuroS\&P)}, pages 372--387. IEEE.

\bibitem[Sartor and Lagioia, 2020]{gdpr}
Sartor, G. and Lagioia, F. (2020).
\newblock {The impact of the General Data Protection Regulation ({GDPR}) on
  artificial intelligence}.
\newblock {\em European Parliamentary Research Service}, pages 1--100.

\bibitem[Sharif et~al., 2016]{sharif2016accessorize}
Sharif, M., Bhagavatula, S., Bauer, L., and Reiter, M.~K. (2016).
\newblock Accessorize to a crime: Real and stealthy attacks on state-of-the-art
  face recognition.
\newblock In {\em Proceedings of the 2016 ACM SIGSAC Conference on Computer and
  Communications Security}, pages 1528--1540. ACM.

\bibitem[Smith and Gal, 2018]{smith2018understanding}
Smith, L. and Gal, Y. (2018).
\newblock Understanding measures of uncertainty for adversarial example
  detection.
\newblock In Globerson, A. and Silva, R., editors, {\em Proceedings of the
  Thirty-Fourth Conference on Uncertainty in Artificial Intelligence, {UAI}
  2018, Monterey, California, USA, August 6-10, 2018}, pages 560--569. {AUAI}
  Press.

\bibitem[Spangher et~al., 2018]{spangher2018actionable}
Spangher, A., Ustun, B., and Liu, Y. (2018).
\newblock Actionable recourse in linear classification.
\newblock In {\em Proceedings of the 5th Workshop on Fairness, Accountability
  and Transparency in Machine Learning}.

\bibitem[Szegedy et~al., 2013]{szegedy2013intriguing}
Szegedy, C., Zaremba, W., Sutskever, I., Bruna, J., Erhan, D., Goodfellow, I.,
  and Fergus, R. (2013).
\newblock Intriguing properties of neural networks.
\newblock {\em arXiv preprint arXiv:1312.6199}.

\bibitem[Tsipras et~al., 2019]{tsipras2018robustness}
Tsipras, D., Santurkar, S., Engstrom, L., Turner, A., and Madry, A. (2019).
\newblock Robustness may be at odds with accuracy.
\newblock In {\em International Conference on Learning Representations}.

\bibitem[Van~Amersfoort et~al., 2020]{van2020uncertainty}
Van~Amersfoort, J., Smith, L., Teh, Y.~W., and Gal, Y. (2020).
\newblock Uncertainty estimation using a single deep deterministic neural
  network.
\newblock In {\em International Conference on Machine Learning}, pages
  9690--9700. PMLR.

\bibitem[Van~Looveren and Klaise, 2019]{van2019interpretable}
Van~Looveren, A. and Klaise, J. (2019).
\newblock Interpretable counterfactual explanations guided by prototypes.
\newblock {\em arXiv preprint arXiv:1907.02584}.

\bibitem[Verma et~al., 2020]{verma2020counterfactual}
Verma, S., Dickerson, J., and Hines, K. (2020).
\newblock {Counterfactual Explanations for Machine Learning: A Review}.
\newblock {\em arXiv preprint arXiv:2010.10596}.

\bibitem[Wachter et~al., 2017]{wachter2017counterfactual}
Wachter, S., Mittelstadt, B., and Russell, C. (2017).
\newblock Counterfactual explanations without opening the black box: Automated
  decisions and the {GDPR}.
\newblock {\em Harv. JL \& Tech.}, 31:841.

\end{thebibliography}

\newpage
\onecolumn
\appendix

\section*{\LARGE SUPPLEMENTARY MATERIAL}

\section{DERIVATIONS} \label{proofs}

Below, we prove Proposition \ref{prop:same_minimum}, which states

\propsameminimum*

We start by introducing the notation, definitions, and our assumptions.

Let  $p_f(y'|x')$ denote the softmax probability assigned to class $y'$  for input $x'$ by the classifier $f$.
Let $\mathcal{X}$  and $\mathcal{Y}$ denote the domains of $x$ and $y$, respectively. We assume $0 \cdot \log(0):=0$; $\log(0) := -\infty$; and $p_f(y|x)>0 , \forall x \in \mathcal{X}, y\in \mathcal{Y}$.

For simplicity, we provide the derivation for a single input $x' \in \mathcal{X}$ with target class $y'$. However, the proof can be easily extended to multiple observations. When generating a CE $x'$ targeted to class $y'$ we minimize the cross-entropy loss, defined as
\begin{equation}
   \ell(x',y') =  - \log p_f(y'| x').
\end{equation}

We observe that the function obtains a minimum at $p_f(y'|x) =1 \text{ and } p_f(y|x)=0 \quad \forall y \in Y \backslash y'$ for which $\ell(x',y')=0$. This is a unique minimum because [1] cross-entropy is bounded below by $0$ and [2] it is monotonically decreasing in $p_f(\cdot)$.

Predictive entropy, $h(x')$, is defined as
\begin{equation}
  h(x') = - \sum_{y \in \mathcal{Y}} p_f(y|x') \log p_f(y|x').
\end{equation}

If $p_f(y'|x) =1, \text{ and } p_f(y|x)=0 \quad \forall y \in Y \backslash y'$, then $h(x')=0$. This is a minimum because predictive entropy is also bounded below by $0$. 

\newpage

\section{COUNTERFACTUAL EXPLANATION GENERATION ALGORITHM} \label{appx:cf_alg}

\begin{algorithm}[H]
\caption{Generating Counterfactuals}
\begin{algorithmic}[1] \label{algorithm_counterfactuals}
\STATE \textbf{Input} original observation $x$; target class $y'$; ensemble of models $ \{f_m \}_{m=1}^M$; maximum number of iterations $N$; minimum confidence of target class $\gamma$; perturbation size $\delta$; maximum number of times each feature is changed $n$; optional: a function that clips the values to a permitted pre-defined range clip.
\STATE \textbf{Output} counterfactual $x'$
\STATE $x' \leftarrow x$
\STATE $c \leftarrow 0$
\STATE Create a $n_p$ vector, where $n_p$, the number of input features, with all values initialized to zero, $P = 0_{n_p}$ This vector will ensure keep track of the number of times each feature has been changed.
\WHILE{$\bar{p}(y'| x') < \gamma$ and $c < N$}
\STATE Compute forward derivative: $S(x',y') = \nabla_{x'} \frac{1}{M}{\ell}(f_m, x', y')$
\STATE Select the most salient feature: $i = \text{argmax}_{i:i \in P, P[i]<n} S(x',y')[i]$
\STATE Update the most salient feature by $\delta$: $x'[i] = x'[i] + \text{sign}(S(x',y')[i]) \cdot \delta$
\STATE Clip the counterfactual so that it stays within the pre-defined range: $x'= \text{clip}(x')$
\STATE $P[i] \leftarrow P[i] + 1$
\STATE $c \leftarrow c+1$
\ENDWHILE
\STATE \textbf{return} $x'$
\end{algorithmic}
\end{algorithm}

\section{ADVERSARIAL TRAINING} \label{appx:adv_training}

Recent work in adversarial literature has linked adversarial robustness to improving model interpretability \citep{tsipras2018robustness}.
Improving adversarial robustness can be achieved through adversarial training, corresponding to minimizing the loss:
\begin{equation}
    \min_{\theta} \E_{p_{\theta}(y|x)} \left[ \max_{\delta \in \Delta} \ell (x+\delta, y| \theta) \right],
\end{equation}
where $\theta$ are the model parameters, $\ell$ is the loss function, $x$ is the original input, $y$ is the original class, $\delta$ is a perturbation, and $\Delta$ is set of possible perturbations.

In practice, adversarial training is often implemented by augmenting the dataset with adversarial examples during training.
These additional images ensure that the model does not focus on noise when learning features for classification.
Thus, the model is more likely to learn features that are not noise, and therefore are more interpretable \citep{tsipras2018robustness, ilyas2019adversarial}.
This means that adversarial training can also be used to improve the performance of models, outside of the adversarial literature setting.

Further,  \cite{chalasani2018concise} show a connection between feature-attribution explanations and adversarial training, finding that it leads to more sparse and stable explanations from [1] an empirical perspective for image data and [2] a theoretical perspective for single-layer models.

Lastly, adversarial training can be used to improve uncertainty estimation. \citet{lakshminarayanan2017simple} show that adversarial training is a computationally efficient solution for smoothing the predictive distribution, and can improve the accuracy and calibration of classifiers in practice.

The aforementioned work motivates the use of adversarial training to generate more interpretable, stable explanations.

\section{EVALUATION} \label{appx:eval_im2}

We omit IM2 evaluation as we found that it failed to pass a sanity check that compares the metric for in- and out-of- distribution images. We perform the check using MNIST, as it made it easier to visually verify test results.

We use MNIST as in-distribution data, and EMNIST as out-of-distribution data. This means that the autoencoders are trained on MNIST images. For the `in-distribution' CEs, we take MNIST training data in the target class -- these images can be considered as `gold standard' CEs. For the `out-of-distribution' CEs, we select a random image from EMNIST. We expect to see a clear different in the IM2 scores for the in- and out-of-distribution data as we are comparing the gold standard CEs with random images from a different dataset.

\begin{figure}[h]
\centering
\includegraphics[width=0.29\linewidth]{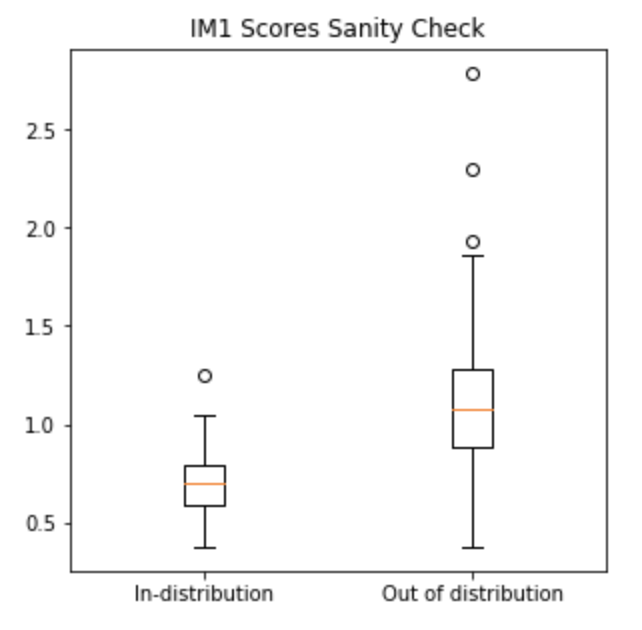}
\includegraphics[width=0.31\linewidth]{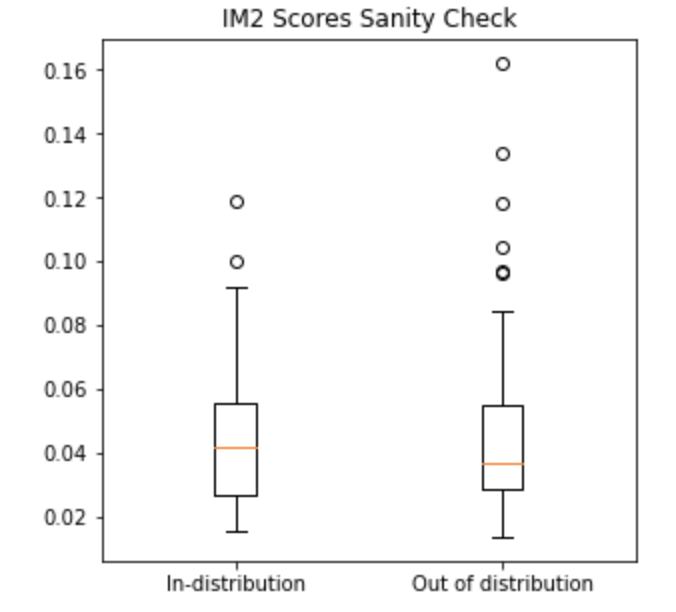}
    \caption{IM1 and IM2 scores for in-distribution data versus out-of-distribution data}
    \label{fig:sanity_check}
\end{figure}

However, the right-hand plot of \Cref{fig:sanity_check} demonstrates that the IM2 scores for out-of-distribution did not differ significantly than IM2 scores of in-distribution data at a $5\%$ significance level.
On the contrary, for IM1, we find a significant difference (measured using a t-test) for in- and out-of distribution IM1 score at a $5\%$ significance level. Visually, the difference can be observed in the left box-plot in Figure \ref{fig:sanity_check}.

\section{EXPERIMENT SETUP} \label{appx:experiment_setup}
Below, we summarize the experiment setup for the different experiments.
All experiments are implemented in PyTorch.
The code repository contains the instructions for reproducing each result and generating example CEs, alongside the details of the environment setup including the versions of dependencies.

\subsection{MNIST}

\paragraph{Dataset Configuration}
For our method and JSMA we normalize the inputs to $[0,1]$.
For \citet{van2019interpretable} we normalize to $[-0.5, 0.5]$, as suggested.
We train the classifier on the training set, and generate CEs on the test set.
We tune the hyperparameters of our method by generating CEs on the training set.
We did not tune the hyperparameters of \citet{van2019interpretable}, as we could select them from the original paper.

For each point in the test set, we select the target class for the explanation by randomly selecting from a set of classes specific to the class of the input image.
This allows us to avoid selecting target classes which are difficult for a particular input class, for example transforming a $6$ into a $2$.
For a complete list of the possible target classes for each input class, see the code release.

\paragraph{Model Architecture} We use a three-layer MLP with $200$, $200$ and $10$ nodes per layer, respectively. The first two layers have ReLU activations and batch normalization after the activation. The final layer has a softmax activation. We use an ensemble of $50$ models.

\paragraph{Optimisation}
We train the network using mini-batches of size $128$ and Adam \citep{kingma2014adam} (with default PyTorch hyper-parameters) to optimize the weights.
We train the model for $50$ epochs.

\paragraph{Our Method: adversarial training} We implement adversarial training as follows: For each iteration, perform a training step a (clean) batch of data $(X, y)$ followed by a training step on an adversarially-augmented batch of data.
To create the latter, we use FGSM \citep{goodfellow2014explaining} to generate adversarial examples $X'$.
The augmented dataset is $(X,y) \cup (X',y)$.
We choose $\varepsilon = 0.15$ as the perturbation is large enough to fool a trained classifier $90 \%$ of the time (and therefore will improve robustness), however does not change the true classification.

\paragraph{Our Method: hyperparameters}
The maximum number of permitted changes is $5$ and the maximum number of iterations is $3$,$920$.
The confidence level, $\gamma$, is set to $0.99$. The perturbation ($\delta$ in the pseudo-code) is set to $0.2$ (which is equal to $1/n$, where $n$ is the number of times each feature can be changed). 

\paragraph{IM1 Evaluation}
We implement $\imone$ evaluation following \citet{van2019interpretable}, and include the configuration here for completeness.
\Cref{table:im1_mnist_aes} shows the architecture of the all class and single class autoencoders.
We train both autoencoders using the mean-squared objective and Adam with a batch size of $128$.
For the all class autoencoder we train for $4$ epochs, for the single class autoencoder we train for $30$ epochs.

\begin{table}[h]
    \begin{center}
    \captionof{table}{
        IM1 MNIST autoencoder architectures.
        Top-down in the table corresponds to input-output in the architecture.
        The hyperparameters are output channels $c$, kernel size $k$, padding $p$, stride $s$, scale $sc$.}
    \vspace{5pt}
    \begin{small}
    \begin{sc}
    \begin{tabular}{l c} \toprule
        \multicolumn{2}{c}{\textbf{All Class}} \\ \midrule
        Layer & Parameters\\ \midrule
        \multicolumn{2}{c}{\textit{Encoder}} \\ \midrule
        2D Convolution & $c=16, k=3, p=1$ \\
        ReLU activation & \\
        2D Convolution & $c=16, k=3, p=1$ \\
        ReLU activation & \\
        Max-pool 2D & $k=2, s=2$\\
        2D Convolution & $c=1, k=3, p=1$ \\ \midrule
        \multicolumn{2}{c}{\textit{Decoder}} \\  \midrule
        2D Convolution & $c=16, k=3, p=1$ \\
        ReLU activation & \\
        Upsample & $sc =2$, mode = ``nearest'' \\
        2D Convolution & $c=16, k=3, p=1$ \\
        ReLU activation & \\
        2D Convolution & $c=1, k=3, p=1$ \\ \bottomrule
    \end{tabular}
    \begin{tabular}{l c} \toprule
        \multicolumn{2}{c}{\textbf{Single Class}} \\ \midrule
        Layer & Parameters\\ \midrule
        \multicolumn{2}{c}{\textit{Encoder}} \\ \midrule
        2D Convolution & $c=16, k=3, p=1$ \\
        ReLU activation & \\
        2D Convolution & $c=16, k=3, p=1$ \\
        ReLU activation & \\
        Max-pool 2D & $k=2, s=2$\\
        2D Convolution & $c=1, k=3, p=1$ \\ \midrule
        \multicolumn{2}{c}{\textit{Decoder}} \\  \midrule
        2D Convolution & $c=16, k=3, p=1$ \\
        ReLU activation & \\
        Upsample & $sc =2$, mode = ``nearest'' \\
        2D Convolution & $c=16, k=3, p=1$ \\
        ReLU activation & \\
        2D Convolution & $c=1, k=3, p=1$ \\ \bottomrule
        \end{tabular}
    \end{sc}
    \end{small}
    \label{table:im1_mnist_aes}
    \end{center}
\end{table}

\paragraph{Results aggregation}
The results in \Cref{table:results} are aggregated as follows:
\begin{itemize}
    \item For each method and dataset pair we choose $100$ points from the test set.
    \item For $10$ random seeds
    \begin{itemize}
        \item Generate a CE for each of the $100$ test points
        \item Compute the mean IM1 score and $L_1$ distance
    \end{itemize}
    \item Compute the mean of the means, and the standard deviation of the means.
\end{itemize}

\paragraph{Configuration of \citet{van2019interpretable}}
We use the implementation released by the authors in the ALIBI library \citep{alibi}, largely in its default configuration as given in the documentation.
We use the same classifier architecture as above.
As the classifier is a white-box model, we follow the ALIBI documentation and use loss function $D$ from \citet{van2019interpretable}.
Thus, the hyperparameter configuration is $c=1$, $\kappa=0$, $\beta=0.1$, $\gamma=100$, $\theta=100$, $\mathrm{max~iterations}=2000$.
As in \citet{van2019interpretable} we use the encoder to find the class prototypes, and following the ALIBI documentation set $K=\textrm{all~instances}$.

\subsection{Wisconsin Breast Cancer Dataset and Boston Housing Dataset} \label{appx:exp_details_wbc}
We generally use the same setup for both the Wisconsin Cancer and the Boston datasets, except where we clearly indicate differences below.

\paragraph{Dataset Configuration}
For our method and JSMA we normalize the inputs to $[0,1]$.
For \citet{van2019interpretable} we standardize the inputs to mean $0$ and standard deviation $1$, as suggested.
We randomly select $70\%$ of each dataset as the train set, $10\%$ as the validation set, and $20\%$ as the test set.
We train the classifier on the training set, tune hyperparameters on the validation set, and generate CEs on the test set.

\paragraph{Model Architecture} We use a three-layer MLP with $80$, $80$ and $2$ nodes per layer, respectively. The first two layers have ReLU activations and batch normalization after the activation. The final layer has a softmax activation. We use an ensemble of $20$ models.

\paragraph{Optimisation} We follow the same setup as for MNIST.

\paragraph{Our Method: adversarial training} We perform adversarial training similarly to MNIST. Contrary to MNIST, each feature has a different scale and distribution. A list of perturbation sizes can be found in the code in the configuration file in demo\_data, after the word perturbation.

\paragraph{Our Method: hyperparameters}
The maximum number of permitted changes is $5$ and the maximum number of iterations is $150$.  The confidence level, $\gamma$, is set to $0.99$. The perturbation sizes ($\delta$ in the pseudo-code) are feature-specific and can be found in the file breast\_cancer\_config.txt in the code. 

\paragraph{IM1 Evaluation}
For the Wisconsin Breast Cancer dataset we follow \citet{van2019interpretable}, and include the configuration here for completeness.
We use the same procedure for the Boston Housing Dataset.
We use the same autoencoder architecture for both the all class and single class autoencoders, and \Cref{table:im1_bc_bh_ae} shows the architecture.
We train both autoencoders using the mean-squared objective and Adam with a batch size of $128$ for $500$ epochs.

\begin{table}[h]
    \centering
    \caption{
        IM1 VAE architecture for the Wisconsin Breast Cancer and Boston Housing datasets.
        The top of the table corresponds to the input of the model, and the bottom the output.
        The hyperparameter $n$ is the number of hidden units.
    }
    \vspace{5pt}
    \begin{small}
    \begin{sc}
    \begin{tabular}{l c}
        \toprule
        Layer & Parameters\\ \midrule
        \multicolumn{2}{c}{\textit{Encoder}} \\ \midrule
        Linear & $n=20$ \\
        ReLU activation \\
        Linear & $n=10$\\
        ReLU activation \\
        Linear & $n=6$\\ \midrule
        \multicolumn{2}{c}{\textit{Decoder}} \\  \midrule
        Linear & $n=6$\\
        ReLU activation \\
        Linear & $n=10$\\
        ReLU activation &\\
        Linear & $n=20$ \\
        \bottomrule
    \end{tabular}
    \end{sc}
    \end{small}
    \label{table:im1_bc_bh_ae}
\end{table}

\paragraph{Results aggregation}
The same as for MNIST.

\paragraph{Configuration of \citet{van2019interpretable}}
We use the implementation released by the authors in the ALIBI library \citep{alibi}, largely in its default configuration as given in the documentation.
We use the same classifier architecture as above.
As the classifier is a white-box model, we follow the ALIBI documentation and use loss function $B$ from \citet{van2019interpretable} for both datasets.
For both datasets we use the kd-tree approach to find the prototypes.
For the Wisconsin Breast Cancer dataset the hyperparameter configuration is $c=1$, $\kappa=0$, $\beta=0.1$, $\gamma=0$, $\theta=100$, $k=1$, $\mathrm{max~iterations}=2000$.
We set $\theta$ by examining the grid search shown in \citet[Figure~7]{van2019interpretable}.
We set $k=1$ as this is the default value in ALIBI.
For the Boston Housing dataset we use the same configuration as for the Wisconsin Breast Cancer dataset.
The exception are $k$ and $\theta$ for which we perform a grid search similar to that run by \citet[Appendix~A]{van2019interpretable} for the Breast Cancer dataset, based on which we choose $k=10$ and $\theta=100$.

We note that the IM1 scores we report for \citet{van2019interpretable} are worse than those in their paper.
They report an IM1 score which is not significantly different to ours.
However, their results are not directly comparable because we use a test set of $100$ points while the original paper uses one of $19$ points.
To try and ensure our results are correct, we took the following steps:
\begin{itemize}
    \item Used the implementation of the generation algorithm provided by the authors
    \item Ensured inputs to the generation algorithm are standardized as in the original paper
    \item Ensured we use the same hyperparameters as in the original paper
    \item Visually examined the CEs generated by the method to ensure they were reasonable
\end{itemize}
We repeated the experiment using a test set of $19$ points rather than $100$, but this also did not reproduce the results in the original paper.

\section{ADDITIONAL FIGURES} \label{appx:add_fig}

\subsection{Uncertainty And Calibration} \label{appx:ensembles}

\subsubsection{Calibration}

\begin{figure}[!h]
\begin{center}
    \includegraphics[width=0.35\linewidth]{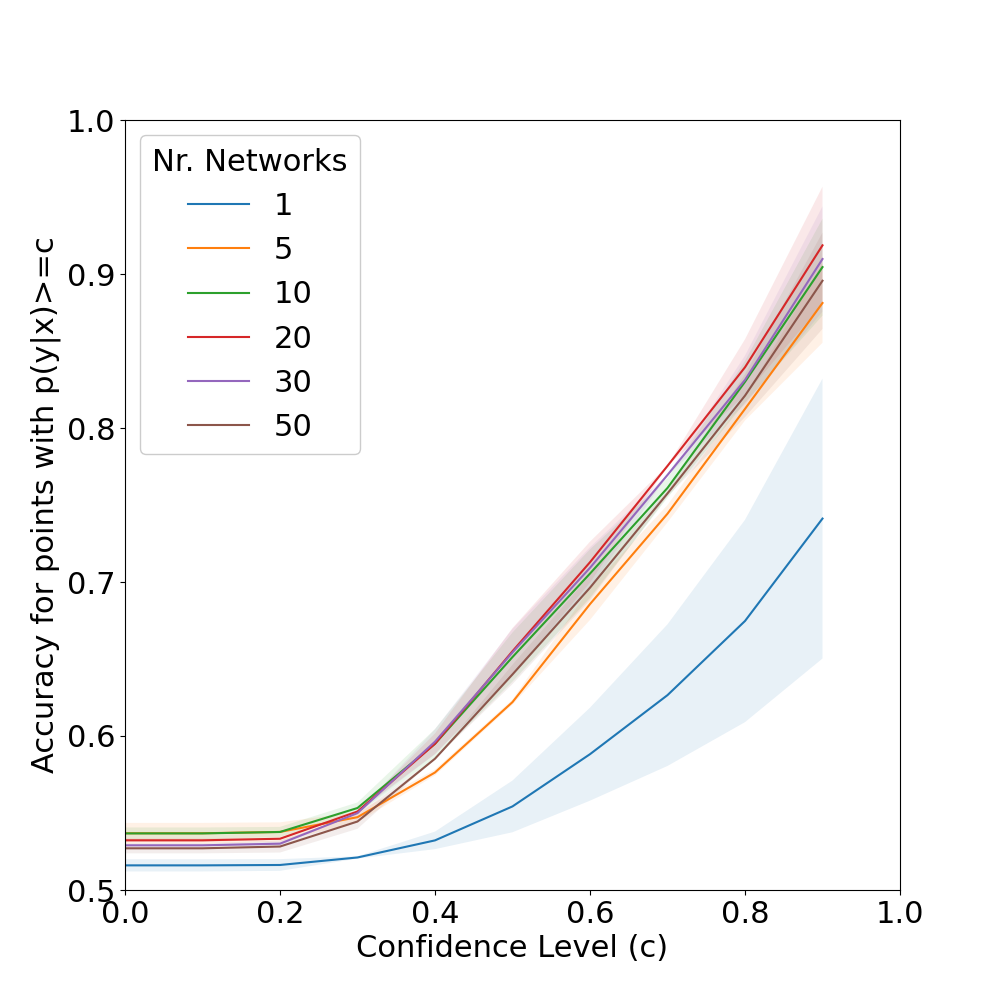}
    \caption{
        Accuracy of the model as a function of confidence, for ensembles containing between $1$ and $50$ models.
        The solid line shows the mean, and the shaded area shows the $95\%$ confidence level over 10 random seeds. A higher line is better. 
    }
    \label{fig:callibration}
\end{center}
\end{figure}

Calibration is important as we assume that our classifier outputs $p(y|x)$.
We investigate the effect of the number of components in the ensemble by performing a similar experiment to \cite[Figure~6]{lakshminarayanan2017simple}, and considering accuracy as a function of confidence.
We train our network (same configuration as in \Cref{appx:experiment_setup}) on MNIST.
We test the network on a dataset formed by combining MNIST and FashionMNIST, where we increase the proportion of the dataset taken from FashionMNIST until the confidence of the classifier falls to the desired level.
Ideally the network will have a low-confidence for incorrect predictions.
Figure \ref{fig:callibration} shows the performance of a single classifer, and ensembles with between $5$ and $50$ components.
We can see that the ensembles perform better than a single classifier as their accuracy is higher. 

\subsubsection{Out-of-Distribution Uncertainty}

Our method is reliant on the quality of the uncertainty estimates offered by the classifier, thus we investigate the effect of ensembling and adversarial training on the uncertainty estimates.
We measure epistemic and aleatoric uncertainty using predictive entropy.
Ideally, we observe a high predictive uncertainty for out-of-distribution data and a low predictive uncertainty for inputs similar to the training data.

\begin{figure}[!h]
\begin{center}
\hspace{0.05\linewidth}
    \begin{minipage}{0.25\textwidth}
    \includegraphics[width=\linewidth]{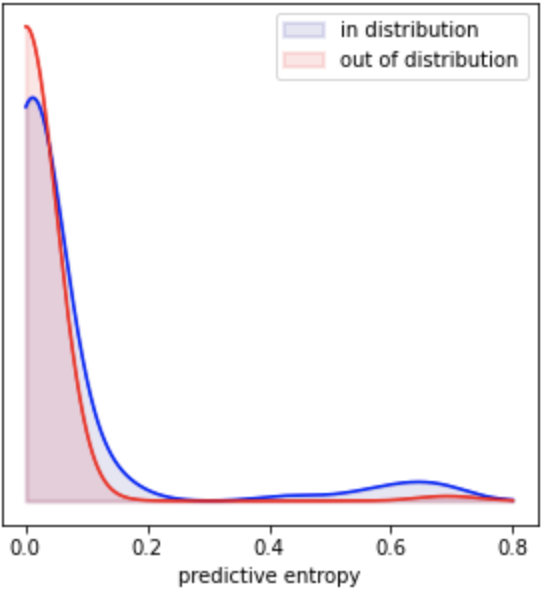}
    \caption*{No Adversarial Training or Ensembling}
    \end{minipage}
    \hspace{1cm}
    \begin{minipage}{0.25\textwidth}
    \includegraphics[ width=\linewidth]{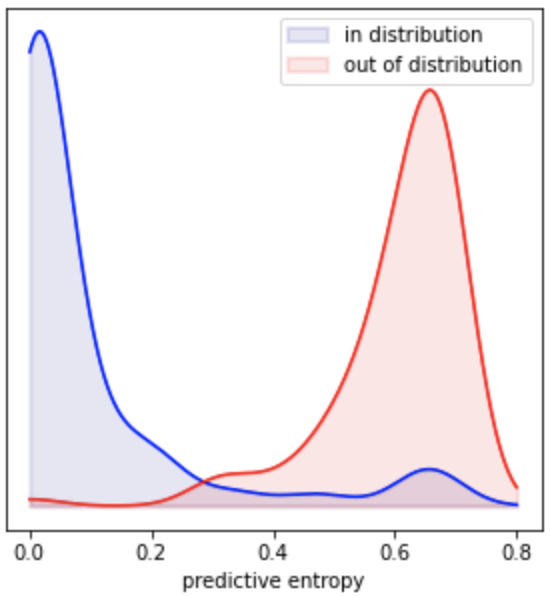}
    \caption*{Adversarial Training and Ensembling}
    \end{minipage}
    \caption{Predictive entropy of the model for in- and out-of distribution inputs. The left plot shows a single model without adversarial training or ensembling. The right plot shows an ensemble of $50$ models with adversarial training.}
    \label{fig:pred_uncertainty}
\end{center}
\end{figure}

We consider in-distribution to be test samples from the Breast Cancer Dataset, and out-of-distribution is tabular data sampled from a normal distribution.
Figure \ref{fig:pred_uncertainty} shows the effect of using adversarial training and ensembling on predictive uncertainty. While a single model cannot distinguish between in- and out-of-distribution inputs, when using ensembling and adversarial training we see a clear separation of the predictive entropy of both sets of inputs.
This goes some way to explain why we see improved performance in the ablation study as we increase the number of models in the ensemble and include adversarial training.

\subsection{Qualitative Analysis}
\begin{figure}[!h]
\centering
\includegraphics[width=0.45\linewidth]{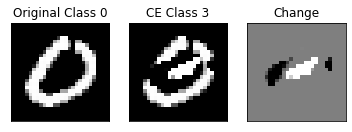}
\hspace{0.05\linewidth}
\includegraphics[width=0.45\linewidth]{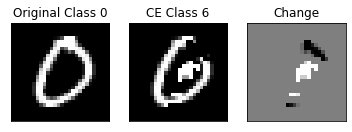}
\\
\includegraphics[width=0.45\linewidth]{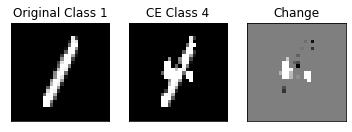}
\hspace{0.05\linewidth}
\includegraphics[width=0.45\linewidth]{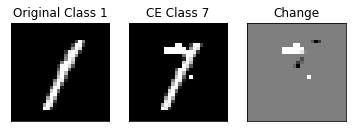}
\\
\includegraphics[width=0.45\linewidth]{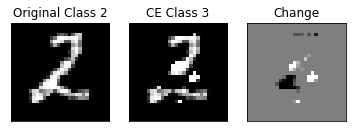}
\hspace{0.05\linewidth}
\includegraphics[width=0.45\linewidth]{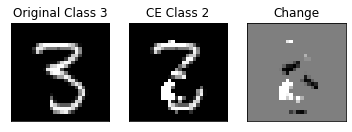}
\\
\includegraphics[width=0.45\linewidth]{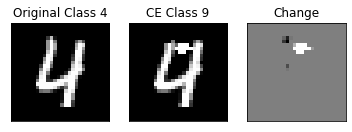}
\hspace{0.05\linewidth}
\includegraphics[width=0.45\linewidth]{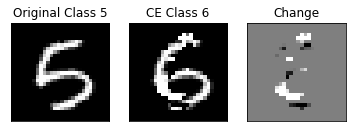}
\\
\includegraphics[width=0.45\linewidth]{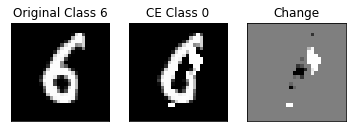}
\hspace{0.05\linewidth}
\includegraphics[width=0.45\linewidth]{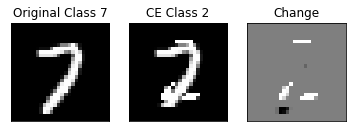}
\\
\includegraphics[width=0.45\linewidth]{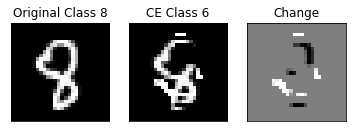}
\hspace{0.05\linewidth}
\includegraphics[width=0.45\linewidth]{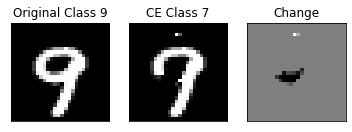}
\caption{Qualitative Examples of Generated CE on MNIST. The first column shows the original image, the second column the counterfactual, and the third column the proposed change (to the original image to generate the counterfactual).}
\label{fig:qual_mnist2}
\end{figure}

Figure \ref{fig:qual_mnist2} shows more qualitative examples of counterfactuals generated by our algorithm on MNIST. The first column shows the original class; the second column shows the counterfactual; the third column shows the proposed change. In the third image in each tuple, black denotes \textit{``painting the pixels black in the original image"}, white denotes \textit{``painting the pixels white in the original image"} and gray denotes \textit{``no change"}. For example, consider the first row in Figure \ref{fig:qual_mnist2}. The goal is to change a $0$ into $3$. Our model proposes:
\begin{itemize}
    \item adding an additional stroke in the middle (shown by the white pixels in the third image),
    \item removing some pixels so that the from the left and right part of the $0$ (shown by the black pixels in the their images).
\end{itemize}

Both changes are required to create a realistic $3$. We observe that, in general, our model can grasp high-level changes that are required. These changes suffice for explanatory purposes. However, our model does not capture stylistic properties, which be seen from the examples in the third row, left side in Figure \ref{fig:qual_mnist2}. Further work is required if we want to employ our method as a generative model. An example of a `failure case' is on left side of the last row -- this example is particularly challenging for our model. However, on a high-level we can see that the algorithm roughly understands the required changes.

\vfill
\newpage

\begin{figure}[!h]
\begin{center}
    \includegraphics[width=0.15\linewidth]{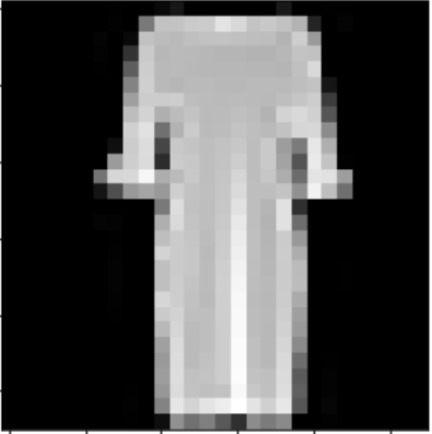}
    \includegraphics[width=0.15\linewidth]{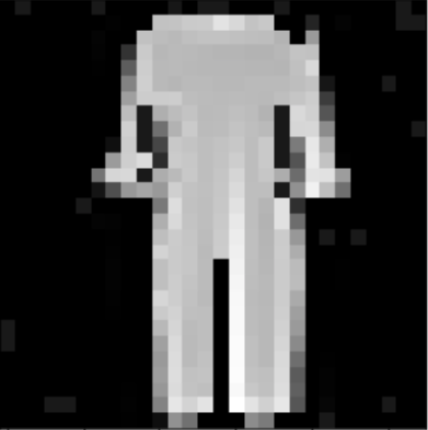}
    \hspace{1cm}
    \includegraphics[width=0.15\linewidth]{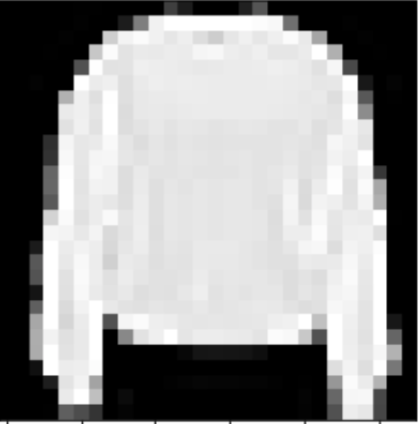}
    \includegraphics[width=0.15\linewidth]{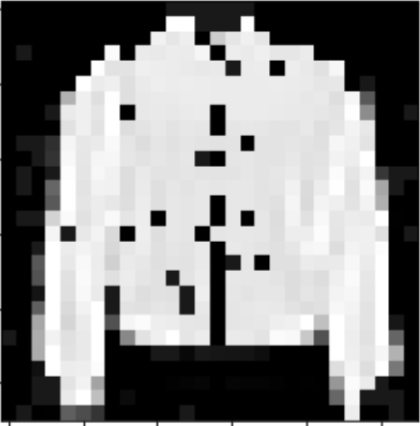}
    \caption{Qualitative Examples of Generated Counterfactuals for FashionMNIST. For each pair, the left image shows the original images: a dress (left pair) and pullover (right). The right images below show generated counterfactuals: pants (left  pair) and a coat (right pair). }
    \label{fig:qual_examplesfmnist}
\end{center}
\end{figure}

Figure \ref{fig:qual_examplesfmnist} shows more qualitative examples of counterfactuals generated by our algorithm on FashionMNIST.  Again, we observe that our model is able to grasp high-level changes required, such as adding a split to the dress or a zipper to the coat.

\newpage
\subsection{Feature Analysis}
The importance of a feature $j$ can be roughly estimated by $\sum_i x_{i,j} -x_{i,j}'$, where $i$ denotes the observation. Using this, we determine which features are most frequently changed. In Figure \ref{fig:size_change_mnist1}, we show examples for MNIST. Here features denote pixels that are changed.

For each triple, the left image shows the average input image from the training dataset -- this represents a prototype of the MNIST digit. The blurriness is caused by the natural variation of the class within the dataset. The second image in each triplet is the average CE generated for the target class shown above the image. The third row shows the most average pixel change, i.e.  $1/n \sum_i x_{i,j} -x_{i,j}'$. In the third image, black denotes \textit{``painting the pixels black in the original image"}, white denotes \textit{``painting the pixels white in the original image"} and gray denotes \textit{``no change"}.

\begin{figure}[h]
\centering
\includegraphics[width=0.45\linewidth]{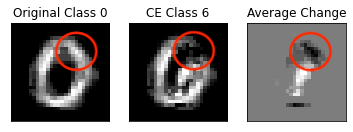}
\hspace{0.05\linewidth}
\includegraphics[width=0.45\linewidth]{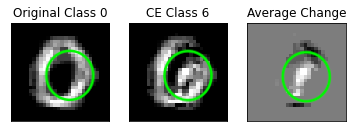}
\caption{Average Feature Perturbation for MNIST. In each tuple: the left image shows the average original input image; the middle image is the average CE; the right image is the average change. }
\label{fig:size_change_mnist_expl}
\end{figure}

\begin{figure}[h]
\centering
\includegraphics[width=0.45\linewidth]{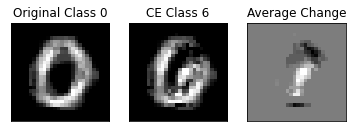}
\hspace{0.05\linewidth}
\includegraphics[width=0.45\linewidth]{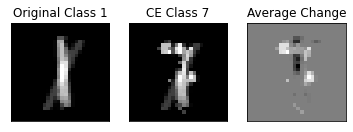} 
\\
\includegraphics[width=0.45\linewidth]{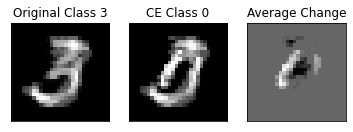}
\hspace{0.05\linewidth}
\includegraphics[width=0.45\linewidth]{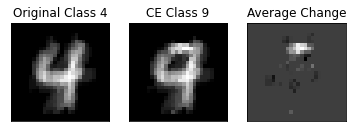}
\\
\includegraphics[width=0.45\linewidth]{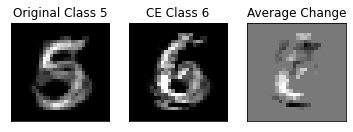}
\hspace{0.05\linewidth}
\includegraphics[width=0.45\linewidth]{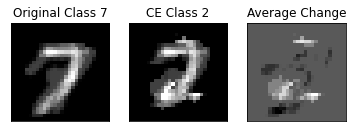}
\caption{Average Feature Perturbation for MNIST. The left image shows the average original input image. The middle image is the average CE. The right image is the average change. }
\label{fig:size_change_mnist1}
\end{figure}

Overall, the proposed changes appear to be realistic.
Let consider a specific example: a counterfactual explanation that changes the predicted class from $0$ to $6$, as shown in the first row of Figure \ref{fig:size_change_mnist1}. In Figure \ref{fig:size_change_mnist_expl}, we highlight the key changes that can be read from Figure \ref{fig:size_change_mnist1}.
The red circles show the parts that have been \textit{removed} from the original input (left image) to create the counterfactual (shown in the middle image). To change a zero into a 6, we need to paint some pixels black at the top right -- these pixels are shown in black in the average change plot (i.e., right image).
The green circles show the parts that have been \textit{added} to the original input (left image) to create the counterfactual (shown in the middle image). To change a zero into a 6, we need to add a white diagonal stroke  --  these pixels are shown in white in the average change plot (i.e., right image).

\begin{figure}
\begin{center}
    \begin{minipage}{0.45\textwidth}
    \includegraphics[width=\linewidth]{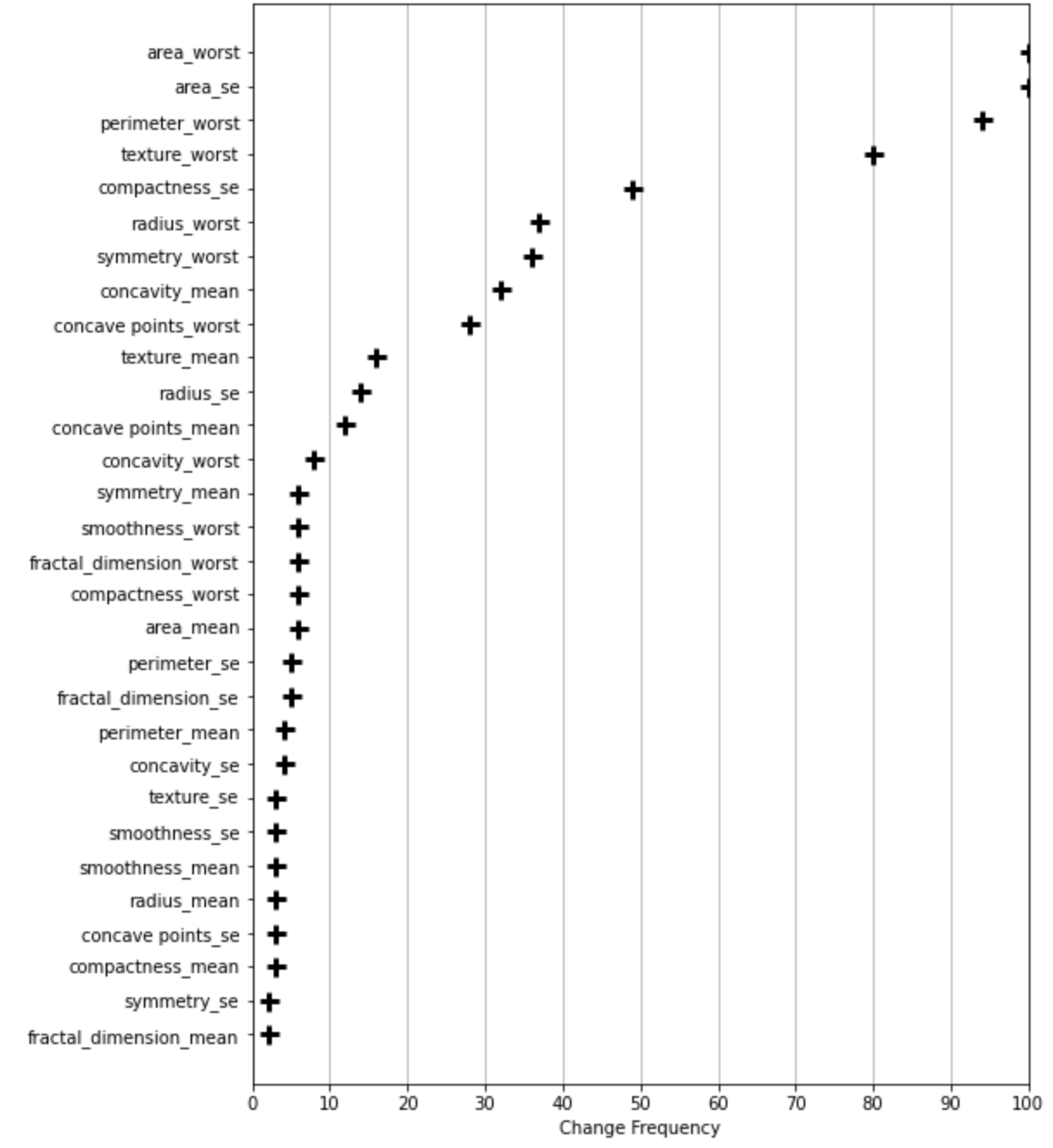}
    \caption{Frequency at which features are changed in counterfactual explanations}
    \label{fig:freq_changed}
    \end{minipage}
    \hspace{0.05\linewidth}
    \begin{minipage}{0.45\textwidth}
    \includegraphics[width=\linewidth]{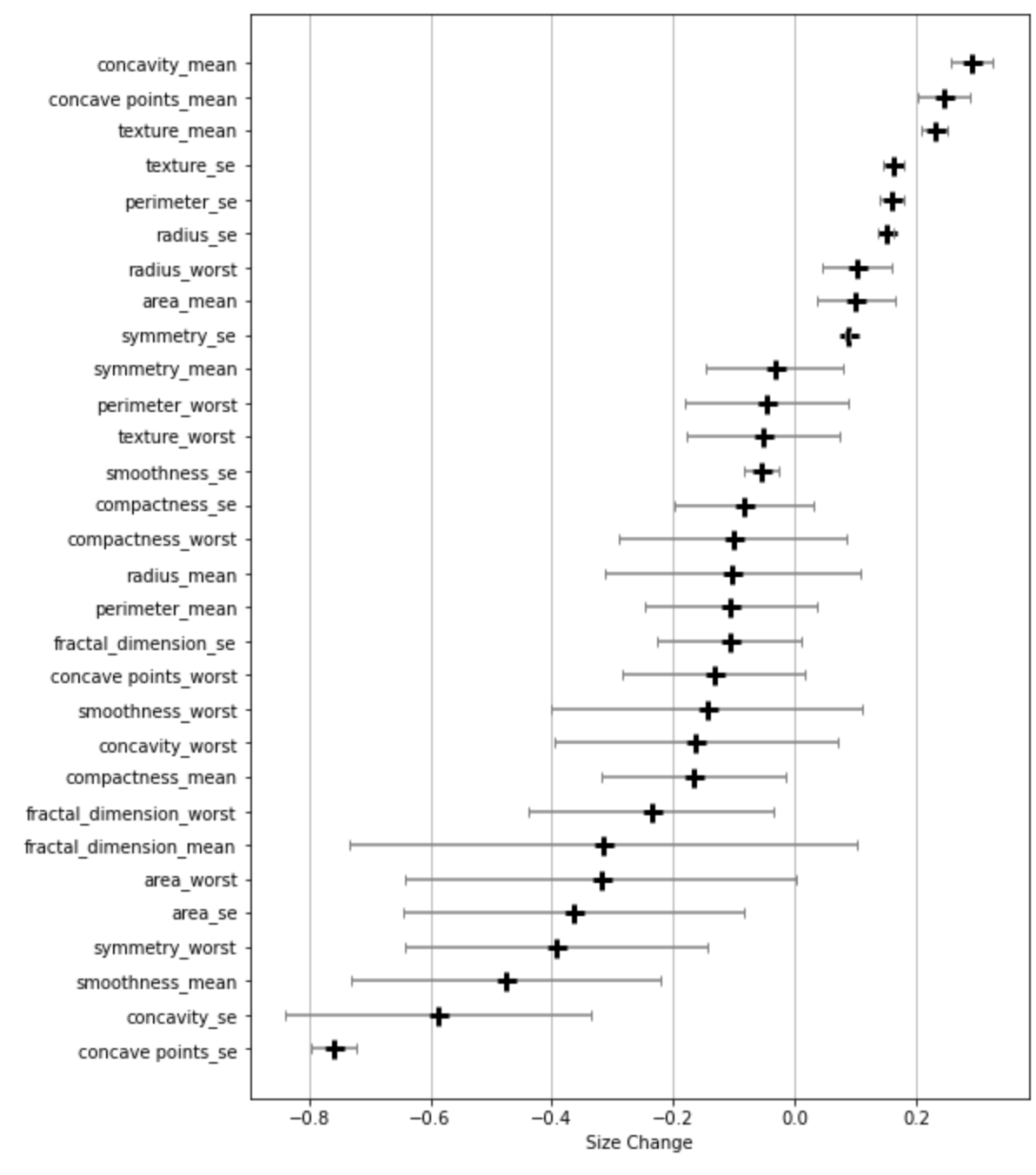}
    \caption{Average Feature Perturbation (of altered features)}
    \label{fig:size_change}
    \end{minipage}
\end{center}
\end{figure}

Figure \ref{fig:freq_changed} shows the most frequently changed cell nuclei properties, and Figure \ref{fig:size_change} shoes the average perturbation size per changed feature. Further interpretation of these results requires expert knowledge, which we intend to look into for future work.

\vfill
\newpage

\end{document}